%% file: ranlp2023.tex
\documentclass[11pt,a4paper]{article}
\usepackage[hyperref]{ranlp2023}
\usepackage{times}
\usepackage{latexsym}

\usepackage{microtype}

\aclfinalcopy 
\usepackage{color}
\usepackage{siunitx}
\usepackage{array}
\usepackage{multirow}
\usepackage{amssymb}
\usepackage{tabularx}

\usepackage{graphicx}
\usepackage{extarrows}
\usepackage{booktabs}
\usepackage{colortbl}

\usepackage{subcaption}

\usepackage{color}
\usepackage{tikz}
\usetikzlibrary{shapes.geometric}
\newcommand{\redmarker}{
\begin{tikzpicture}
    \node[diamond,draw,scale=0.4,fill=red] (d) at (0,0) {};
\end{tikzpicture}
}

\usepackage{standalone} 
\usepackage{tikz}

\usetikzlibrary{fadings}
\usetikzlibrary{patterns}
\usetikzlibrary{shadows.blur}
\usetikzlibrary{shapes}

\usepackage{epsfig}

\usepackage{psfrag}

\title{Unimodal Intermediate Training for \\Multimodal Meme Sentiment Classification}

\author{Muzhaffar Hazman$^{1}$,
Susan McKeever$^{2}$, \and 
Josephine Griffith$^{1}$\\ 
$^{1}$University of Galway, Ireland\\
$^{2}$Technological University Dublin, Ireland \\
\texttt{\{m.hazman1,josephine.griffith\}@universityofgalway.ie}\\
\texttt{susan.mckeever@TUDublin.ie}}

\date{}

\begin{document}
\maketitle
\begin{abstract}
Internet Memes remain a challenging form of user-generated content for automated sentiment classification. The availability of labelled memes is a barrier to developing sentiment classifiers of multimodal memes. To address the shortage of labelled memes, we propose to supplement the training of a multimodal meme classifier with unimodal (image-only and text-only) data. In this work, we present a novel variant of supervised intermediate training that uses relatively abundant sentiment-labelled unimodal data. Our results show a statistically significant performance improvement from the incorporation of unimodal text data. Furthermore, we show that the training set of labelled memes can be reduced by 40\% without reducing the performance of the downstream model.

\end{abstract}

\section{Introduction}

As Internet Memes (or just ``\textbf{memes}") become increasingly popular and commonplace across digital communities worldwide, research interest to extend natural language classification tasks, such as sentiment classification, hate speech detection, and sarcasm detection, to these multimodal units of expression has increased. However, state-of-the-art multimodal meme sentiment classifiers significantly underperform contemporary text sentiment classifiers and image sentiment classifiers. Without accurate and reliable methods to identify the sentiment of  multimodal memes, social media sentiment analysis methods must either ignore or inaccurately infer opinions expressed via memes. As memes continue to be a mainstay in online discourse, our ability to infer the meaning they convey becomes increasingly pertinent \cite{memo1_report,memo3_data}.

Achieving similar levels of sentiment classification performance on memes as on unimodal content remains a challenge. In addition to its multimodal nature, multimodal meme classifiers must discern sentiment from culturally specific inputs that comprise brief texts, cultural references, and visual symbolism \cite{memes_capital}. Although various approaches have been used to extract information from each modality (text and image) recent works have highlighted that meme classifiers must also recognise the various forms of interactions between these two modalities \cite{zhu,aomd,hazman_aics}. 

Current approaches to training meme classifiers are dependent on datasets of labelled memes \cite{hateful_data,memo1_report,suryawanshi-etal-2020-dataset,memo2_report,memo3_data} containing sufficient samples to train classifiers to extract relevant features from each modality and relevant cross-modal interactions. Relative to the complexity of the task, the current availability of labelled memes still poses a problem, as many current works call for more data \cite{zhu,hateful_data,sharma-etal-2022-domain}.

Worse still, memes are hard to label. The complexity and culture dependence of memes \cite{shifman_identity} cause the Subjective Perception Problem \cite{memo1_report}, where varying familiarity and emotional reaction to the contents of a meme from each annotator causes different ground-truth labels. Second, memes often contain copyright-protected visual elements taken from other popular media \cite{meme_intertext}, raising concerns when publishing datasets. This required \citet{hateful_data} to manually reconstruct each meme in their dataset using licenced images, significantly increasing the annotation effort. Furthermore, the visual elements that comprise a given meme often emerge as a sudden trend that rapidly spreads through online communities \cite{meme_2011,shifman_photos}, quickly introducing new semantically rich visual symbols into the common meme parlance, which carried little meaning before \cite{shifman_quiddity}. Taken together, these characteristics make the labelling of memes particularly challenging and costly.

In seeking more data-efficient methods to train meme sentiment classifiers, our work attempts to leverage the relatively abundant unimodal sentiment-labelled data, i.e. sentiment analysis datasets with image-only and text-only samples. We do so using \citeauthor{phang2019sentence}'s \citeyearpar{phang2019sentence} \textbf{S}upplementary \textbf{T}raining on \textbf{I}ntermediate \textbf{L}abeled-data \textbf{T}asks (\textbf{STILT}) which addresses the low performance often encountered when finetuning pretrained text encoders to data-scarce Natural Language Understanding (NLU) tasks. \citeauthor{phang2019sentence}'s STILT approach entails three steps:

\begin{enumerate}
    \item Load pretrained weights into a classifier model.
    \item Finetune the model on a supervised learning task for which data is easily available (the \textbf{intermediate task}).
    \item Finetune the model on a data-scarce task (the \textbf{target task}) that is distinct to the intermediate task.
\end{enumerate}

STILT has been shown to improve the performance of various models in a variety of text-only target tasks \cite{poth-etal-2021-pre,wang-etal-2019-tell}. Furthermore, \citet{pruksachatkun-etal-2020-intermediate} observed that STILT is particularly effective in target tasks in NLU with smaller datasets, e.g. \textit{ WiC} \cite{pilehvar-camacho-collados-2019-wic} and \textit{BoolQ} \cite{clark-etal-2019-boolq}. However, they also showed that the performance benefits of this approach are inconsistent and depend on choosing \textit{appropriate} intermediate tasks for any given target task. In some cases, intermediate training was found to be detrimental to target task performance; which \citet{pruksachatkun-etal-2020-intermediate} attributed to differences between the required ``\textbf{syntactic and semantic} \textbf{\textit{skills}}" needed for each intermediate and target task pair. However, STILT has not yet been tested in a configuration in which intermediate and target tasks have different input modalities.

\begin{table*}[]
\small

\centering

\caption{Sample (a) multimodal memes \cite{memo2_data}, (b) unimodal images \cite{crowdflower_2016}, and (c) unimodal text \cite{potts-etal-2021-dynasent} from the datasets used. Unimodal images and texts of neutral sentiment not pictured here.   \label{tab_samples}}
\arrayrulecolor{lightgray}
\begin{tabularx}{\textwidth}{m{0.25in}>{\centering\arraybackslash}X|>{\centering\arraybackslash}X|>{\centering\arraybackslash}X|>{\centering\arraybackslash}X|>{\centering\arraybackslash}X|>{\centering\arraybackslash}X|>{\centering\arraybackslash}X}
\arrayrulecolor{black}
\hline
\arrayrulecolor{lightgray}

  & \multicolumn{3}{c|}{\textbf{(a) Meme}} & \multicolumn{2}{c|}{\textbf{(b) Image}} & \multicolumn{2}{c}{\textbf{(c) Text}}
\\
& \textbf{(i)} & \textbf{(ii)} & \textbf{(iii)} & \textbf{(i)} & \textbf{(ii)} & \textbf{(i)} & \textbf{(ii)}\\

\arrayrulecolor{black}
\hline

 \shortstack{\textbf{Input} \\\textbf{Image}} &  \includegraphics[width = \hsize]{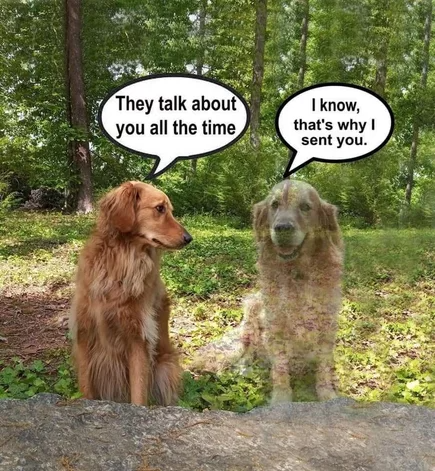} &  \includegraphics[width = \hsize]{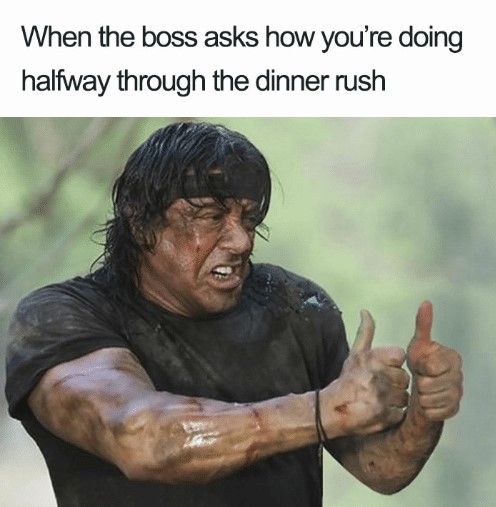} &  \includegraphics[width = \hsize]{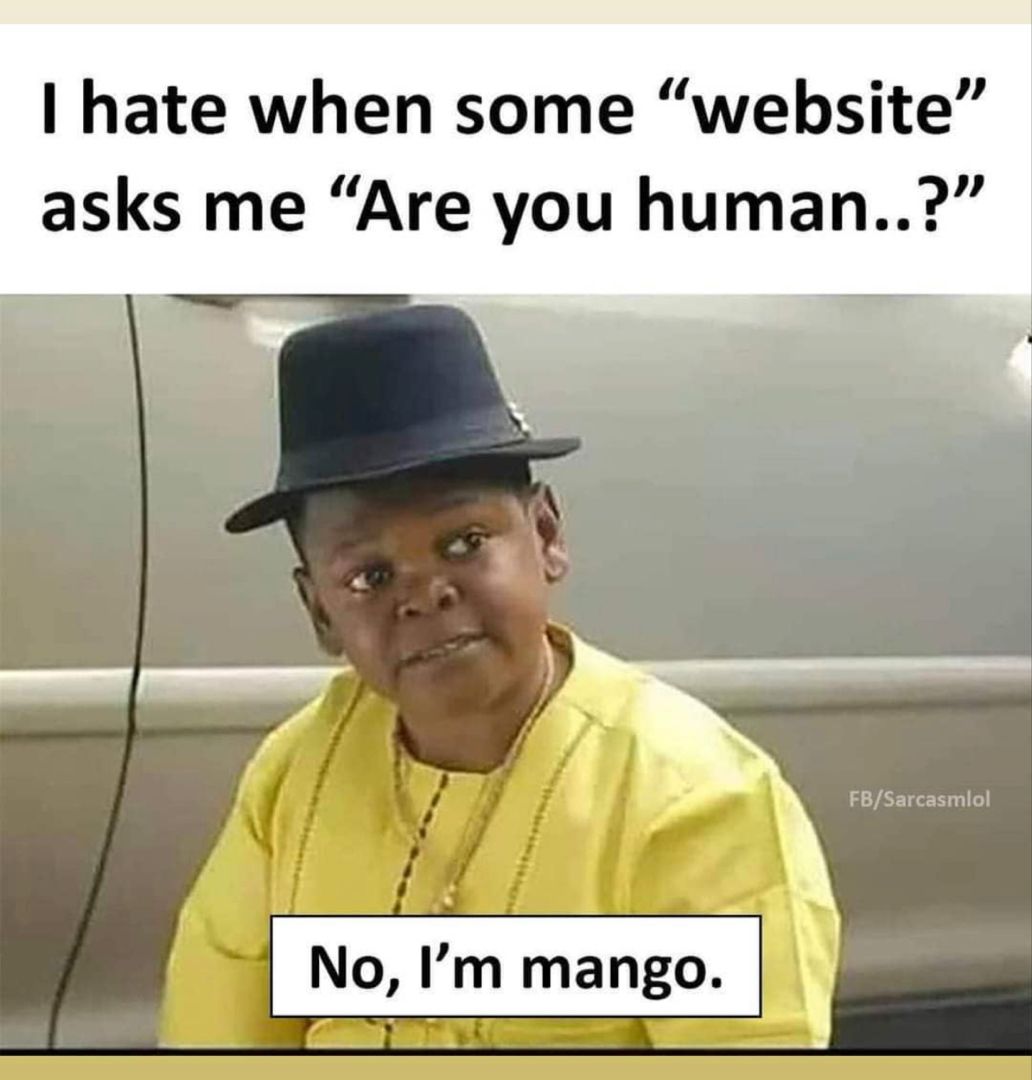} &  \includegraphics[width = \hsize]{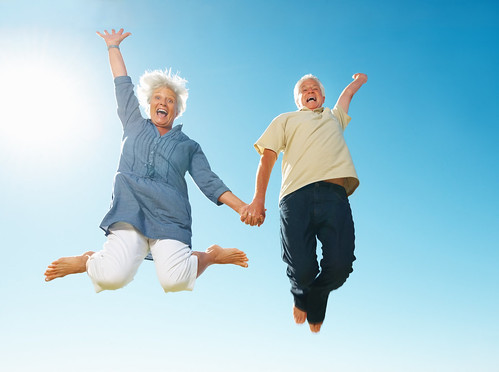} &  \includegraphics[width = \hsize]{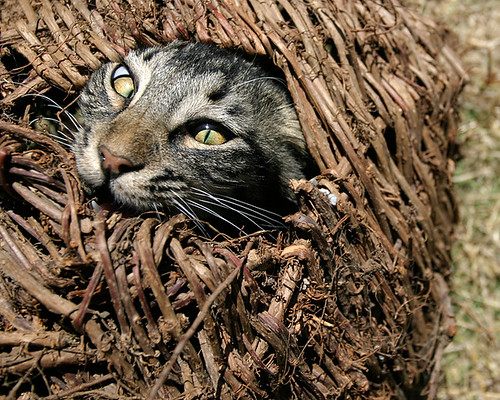}
 & --
 & --\\

\textbf{Input Text} & \scriptsize \texttt{they talk about you all the time i know thats why i sent you} & 
\scriptsize \texttt{when the boss asks how youre doing halfway through the dinner rush } & 
\scriptsize \texttt{i hate when some website asks me are you human  no im mango } & 
--
& --
& \scriptsize \texttt{I tried a new place. I can't wait to return and try more.}
& \scriptsize \texttt{My wife was disappointed.}

\\
\arrayrulecolor{lightgray}\hline
\textbf{Label} & Positive & Neutral & Negative & Positive & Negative & Positive & Negative\\

\arrayrulecolor{black}\hline

\end{tabularx}

\end{table*}

 Although only considering the text or image of a meme in isolation does not convey its entire meaning \cite{hateful_data}, we suspect that unimodal sentiment data may help incorporate \textit{skills} relevant to discern the sentiment of memes. By proposing a novel variant of STILT that uses unimodal sentiment analysis data as an intermediate task in training a multimodal meme sentiment classifier, we answer the following questions:

\paragraph{RQ1:} Does supplementing the training of a multimodal meme classifier with unimodal sentiment data significantly improve its performance?

\begin{figure}[t]
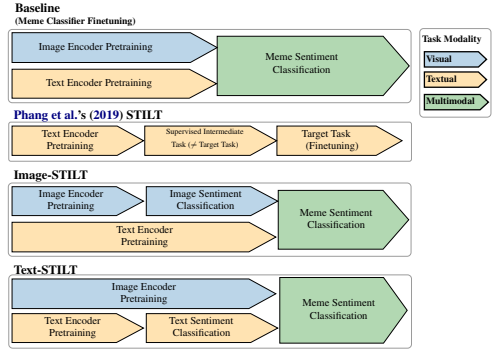

    \centering
    \includestandalone[width=0.83\columnwidth]{figures/diag_train_wide}
    \caption{Training tasks in Baseline, \citeauthor{phang2019sentence}'s \citeyearpar{phang2019sentence} STILT, and our proposed Image-STILT and Text-STILT approaches. \label{diag_train}}
    
\end{figure}

We separately tested our proposed approach with image-only and text-only 3-class sentiment data (creating \textbf{Image-STILT} and \textbf{Text-STILT}, respectively) as illustrated in Figure \ref{diag_train}). If either proves effective, we additionally answer:
\paragraph{RQ2:} With unimodal STILT, to what extent can we reduce the amount of labelled memes whilst preserving the performance of a meme sentiment classifier?

\section{Related Works}
\subsection{Meme Affective Classifiers}

Meme sentiment classifiers fall within the broader category of meme affective classifiers, which can be defined as multimodal deep learning models trained to classify memes by a given affect dimension, including sentiment polarity, offensiveness, motivationality, sarcasm \cite{memo1_report,memo2_report,memo3_data}, hate speech \cite{hateful_data}, and trolling behaviour \cite{suryawanshi-etal-2020-dataset}.
Based on the majority of state-of-the-art meme classifiers, the current literature suggests that these different tasks do not require architecturally distinct solutions \cite{hazman_aics}. Broadly, two general architectural approaches exist among multimodal meme affective classifiers: first, multi-encoder models that use multiple pretrained unimodal encoders which are then fused prior to classification -- numerous examples are summarised by \citeauthor{memo1_report} \citeyearpar{memo1_report} and \citeauthor{memo2_report} \citeyearpar{memo2_report}. These models use both a text encoder and an image encoder that were each trained in unimodal self-supervised and unsupervised tasks such as BERT or SentenceTransformer for text, and VGG-19 or RESNET50 for images. In contrast, single-encoder models are based on a pretrained multimodal vision-and-language model, most often a transformer that has been pretrained on multimodal tasks and accepts both modalities as a single input. The single-encoder approach \cite{hateful_muennighoff,zhu} reuses models that have been pretrained on multimodal tasks such as VL-BERT, UNITER, ERNIE-ViL, DeVLBERT and VisualBERT. There is little empirical evidence to show that one architectural approach consistently outperforms the other in the various meme classification tasks.

Typically, both multi- and single-encoder architectures use transfer learning by finetuning pretrained models on a dataset of labelled memes. While pretraining is often assumed to yield performance benefits for meme classification tasks, this has not been exhaustively proven, especially when viewed relative to studies in image- and text-only tasks \cite{jiang2022transferability}. Multimodally pretrained baseline models for the Hateful Memes dataset \cite{hateful_data} outperformed their unimodally pretrained counterparts. \citet{suryawanshi-etal-2020-dataset} showed that the use of pretrained weights did not consistently provide performance benefits to their image-only classifiers of trolling behaviour in Tamil code-mixed memes. Although the use of pretrained encoders is common amongst meme sentiment classifiers \cite{sharma-etal-2022-domain,blue,mhameme, memo1_report,memo2_report}, there is little evidence as to whether pretrained representations are suitable for the downstream task or if an encoder's performance in classifying unimodal input transfers to classifying multimodal memes.

Beyond using pretrained image and text encoders, several recent works have attempted to incorporate external knowledge into meme classifiers. Some employed additional encoders to augment the image modality representation such as human faces \cite{zhu,hazman_aics}, while others have incorporated image attributes (including entity recognition via a large knowledge base) \cite{momenta}, cross-modal positional information \cite{aomd,hazman_aics}, social media interactions \cite{aomd}, and image captioning \cite{blaier-etal-2021-caption}. To our knowledge, no published attempts have been made to directly incorporate unimodal sentiment analysis data into a multimodal meme classifier.

\subsection{Supplementary Training of Meme Classifiers}\label{litt_supp}
Several recent works addressed the lack of labelled multimodal memes by incorporating additional non-meme data. \citet{sharma-etal-2022-domain} presents two self-supervised representation learning approaches to learn the ``semantically rich cross-modal feature[s]'' needed in various meme affective classification tasks. They finetuned an image and a text encoder on image-with-caption tweets before fitting these representations on to several multimodal meme classification tasks including sentiment, sarcasm, humour, offence, motivationality, and hate speech. These approaches showed performance improvement on some, but not all, tasks. In some cases, their approach underperfomed in comparison to the more typical supervised finetuning approaches. Crucially, since the authors did not compare their performance to that of the same architecture without the self-supervised step, isolating performance gains directly attributable to this step is challenging. Furthermore, while the authors reported multiple tasks where their approach performed best while training on only 1\% of the available memes, their included training curves imply that these performance figures were selected at the point of maximum performance on the testing set during training. This differs from the typical approach of early stopping based on performance on a separately defined validation set, which hinders direct comparisons to competing solutions.

\citet{blue} proposed a multitask learning approach that simultaneously trained a classifier on different meme classification tasks -- sentiment, sarcasm, humour, offence, motivationality -- for the same meme inputs. Their results showed that multitask learning underperformed in the binary detection of humour, sarcasm, and offensiveness. This approach was found to be only effective in predicting the intensity of sarcasm and offensiveness of a meme. However, in sentiment classification, this multitask approach showed inconsistent results. Although multitask learning did not improve the performance of their text-only classifier, their multitask multimodal classifier offers the best reported results on the Memotion 2.0 sentiment classification task to date. 

To the best of our knowledge, only one previous work used unimodal inputs to supplement training of multimodal meme classifiers. \citeauthor{suryawanshi-etal-2020-dataset}'s \citeyearpar{suryawanshi-etal-2020-dataset} initial benchmarking of the TamilMemes dataset showed that the inclusion of unimodal images improved the performance of their ResNet-based image-only model in detecting trolling behaviour in Tamil memes. The authors augmented their dataset of memes with images collected from Flickr; by assigning these images as not containing trolling language. They found that this augmentation with 1,000 non-meme images decreased the performance of their classifier. With 30,000 images, their classifier performed identically to one that only used pretrained weights and supervised training on memes; both were outperformed by their model that did not use either pretrained weights or data augmentations.

Existing supplementary approaches to improve meme classification performance have shown mixed results. Notably, the observations made in these works were measured only once and were not accompanied by statistical significance tests, necessitating caution when drawing conclusions on their effectiveness.

\section{Methodology}

To address our research questions, we chose the 3-class sentiment polarity of multimodal memes as our target task as defined by \citeauthor{memo2_data} \citeyearpar{memo2_data} for our chosen dataset. Our experimental approach revolves around comparing the performance of a multimodal classifier trained only on memes (our \textbf{Baseline}) and those trained first on unimodal image or text data (our \textbf{Image-STILT} and \textbf{Text-STILT} models, respectively) before being trained on memes. These models are architecturally identical to each other, all trained in the Memotion 2.0 training set and tested against the Memotion 2.0 testing set to isolate the effect of unimodal intermediate training on meme sentiment classification performance. The results of the performance of the model are measured using the weighted F1-score, as defined by the authors of the selected meme dataset \cite{sharma-etal-2022-domain}. A detailed description of this metric is available in Appendix B.

\subsection{Model Architecture}\label{meth-arch}

\begin{figure}[]
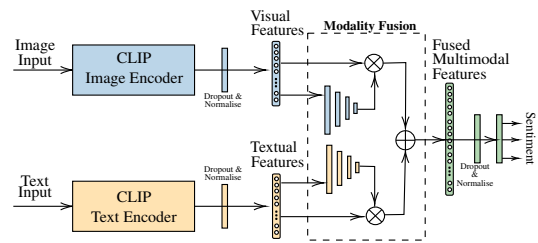

    \centering
    \includestandalone[width=0.9\columnwidth]{figures/diag_arch}
    \caption{Our model architecture. Source: Adapted from \cite{hazman_aics}. \label{diag_arch}}

\end{figure}
As this work does not seek to propose a new meme classifier architecture, we heavily base our model on one found in literature: the \texttt{Baseline} model proposed by \citeauthor{hazman_aics} \citeyearpar{hazman_aics}. Per this previous work, we also use the image and text encoders from OpenAI CLIP \cite{clip} to represent each modality, respectively, and the same modality fusion weighting mechanism they had used. However, we added dropout and batch normalisation after encoding each modality and the fusion of these encodings, which were helpful in preventing overfitting. Figure \ref{diag_arch} illustrates our architecture and a detailed description is presented in Appendix C.

\subsection{Datasets \label{meth-data}}

\arrayrulecolor{lightgray}
\begin{table}[b]
\small
    \centering
    \caption{Meme, Image and Text sample counts in the Memotion 2.0 \cite{memo2_data}, Crowdflower \cite{crowdflower_2016}, DynaSent \cite{potts-etal-2021-dynasent}, respectively.}
    \label{tab_class}
    \begin{tabularx}{\columnwidth}{p{0.65in}>{\arraybackslash}X|>{\arraybackslash}X>{\arraybackslash}X>{\arraybackslash}X>{\arraybackslash}X}
    \arrayrulecolor{black}
    \hline
    \multirow{2}{*}{\textbf{Dataset }} & \multirow{2}{*}& \multicolumn{4}{c}{\textbf{Samples}}\\
    
     & & \textbf{Pos} & \textbf{Neu} & \textbf{Neg} & \textbf{Total}\\
     \hline \\[-7pt]
     \multirow{3}{*}{\textbf{Memotion 2.0}} &  \textbf{Train} & 1,517 & 584 & 172 & 7,000\\
     & \textbf{Val} & 325 & 975 & 200 & 1,500\\
     & \textbf{Test} & 78 & 971 & 451 & 1,500\\
     \arrayrulecolor{lightgray}\hline
    \textbf{Crowdflower}  &  & 5,313 & 1,259 & 1,227 & 7,799\\
    \hline
    \textbf{DynaSent} &  & 6,038  & 5,782 & 4,579 & 16,399\\

    \arrayrulecolor{black}\hline
    
    \end{tabularx}
    
\end{table}

\paragraph{Multimodal Memes:}\label{meth-data-memes}
This work uses sentiment-labelled multimodal memes from the Memotion 2.0 \cite{memo2_data} benchmark dataset as our target task. We did not use the earlier \cite{memo1_report} and later \cite{memo3_data} iterations of this dataset as the former did not provide a validation set and the latter focused on code-mixed languages. Each sample in this meme dataset comprises a meme collected from the web that was then labelled by multiple annotators as conveying either a Positive, Negative or Neutral sentiment. For each meme sample, the dataset presents an image file and a string of the text that was extracted using OCR with manual validation. 

To assess the effectiveness of our approach on various amounts of labelled memes available for training, that is, to answer RQ2, we defined fractional training datasets by randomly sampling the memes training set at the following fractions: 5, 10, 20, 30, 40, 50, 60, 70, and 80\%. For each random restart, we repeat this sampling to account for variance in model performance attributable to training data selection. Where matched pairs are needed for hypothesis testing (see Section \ref{meth-rq2}.RQ2 below), we do not resample between training Baseline, Image-STILT and Text-STILT models. To prevent the models from converging into a model that predicts only the most prevalent class in the training set, we balance the classes in these fractional datasets by applying weights inverse of the class distribution during sampling without replacement. 

\paragraph{Unimodal Images and Texts:}\label{meth-data-uni}
For unimodal intermediate training, we used two unimodal datasets: Crowdflower \cite{crowdflower_2016} for unimodal images, and DynaSent \cite{potts-etal-2021-dynasent} for unimodal text. Both datasets comprise crowdsourced samples collected from social networking sites, and both contain crowd-annotated 3-class sentiment labels\footnote{CrowdFlower's \texttt{Highly Negative} and \texttt{Highly Positive} are treated as \texttt{Negative} and \texttt{Positive}.}. We included all images from the CrowdFlower dataset that we were able to fetch via the provided URLs; not all samples were retrievable. The summaries of, and examples from, these datasets are presented in Tables \ref{tab_class} and \ref{tab_samples}, respectively.

\subsection{Training}

\paragraph{Baseline:} For each run, the model is initialised by loading pretrained weights for the encoders and randomly initialising the weights in the fusion mechanism. For our Baseline approach, the model is trained on the Memotion 2.0 training set, with early stopping at the point of minimum loss on the validation set, and evaluated against the testing set. We maintain the dataset splits defined by \citet{memo2_data}. 
\paragraph{Unimodal STILTs -- Image-STILT and Text-STILT:}
In our proposed approaches, the initialisation of the model is the same as for Baseline and is followed by training the model on a selected unimodal dataset while freezing the encoder of the other modality, that is, the text encoder is frozen while training on unimodal images in Image-STILT and vice versa. Unimodal training ends with early stopping based on the model's performance on the Memotion 2.0 validation set. This model is then trained and tested on the Memotion 2.0 training and testing sets, respectively, as was done in the Baseline approach. Hyperparameters used when training on the Memotion 2.0 dataset are kept constant across all models (see Appendix A). 
\subsection{Experimental Approach}

\arrayrulecolor{lightgray}
\begin{table*}[]
\small

\centering

\caption{Example unimodal images and multimodal memes showing distinct visual symbols. \label{tab_vis}}
\begin{tabularx}{\textwidth}{p{0.7in} |  >{\centering\arraybackslash}X| >{\centering\arraybackslash}X|>{\centering\arraybackslash}X |>{\centering\arraybackslash}X}
 \arrayrulecolor{black}
 
 \hline
\arrayrulecolor{lightgray}
 &  \multicolumn{2}{c|}{\textbf{Image}} & \multicolumn{2}{c}{\textbf{Meme}}\\
 & \textbf{(a)} & \textbf{(b)} & \textbf{(c)} & \textbf{(d)} \\
\arrayrulecolor{black}\hline
&&&&
\\

 &  \includegraphics[width = \hsize]{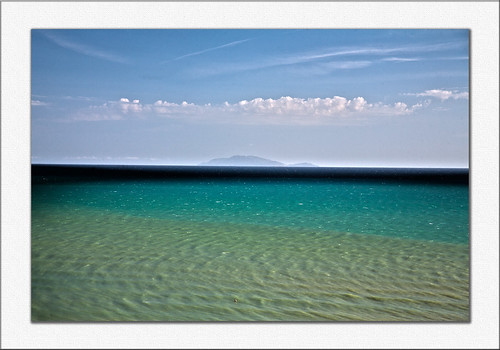} &  \includegraphics[width = \hsize]{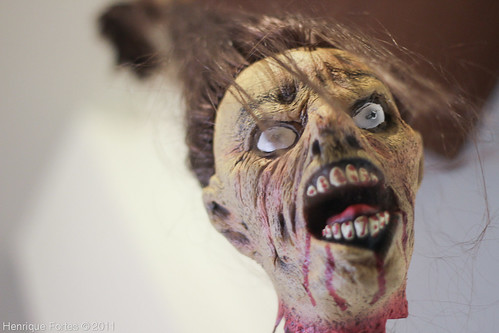} &  \includegraphics[width = 0.9\hsize]{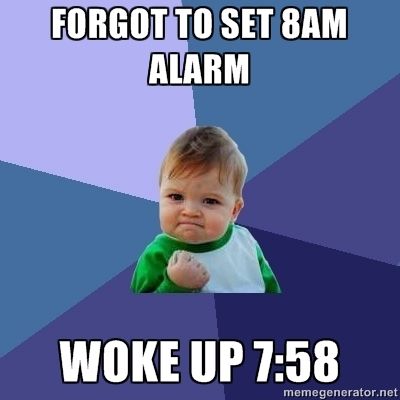} &  \includegraphics[width =0.75\hsize]{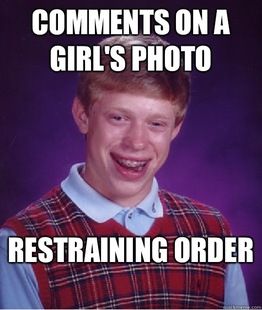}  \\

\arrayrulecolor{lightgray}\hline
\textbf{Sentiment} & Positive & Negative &  Positive & Negative \\[0.1cm]

 \arrayrulecolor{black}
\hline
\end{tabularx}

\end{table*}

\paragraph{RQ1:} To establish whether Image-STILT or Text-STILT offers a statistically significant performance improvement over Baseline, we employ the Wilcoxon Signed-Rank test. The null hypothesis in each case is that there is no significant performance difference between our Baseline approach and Image-STILT or Text-STILT, respectively. We ran 10 random-restarts for each approach: Baseline, Image-STILT, and Text-STILT. All models were trained on all memes from the Memotion 2.0 \cite{memo2_data} training set. Separate tests were conducted for (1) Baseline vs. Image-STILT and (2) Baseline vs. Text-STILT; resulting in a total of 10 pairs each for hypothesis testing. 

\paragraph{RQ2: \label{meth-rq2}} To characterise the performance benefits of Image-STILT or Text-STILT with limited availability of labelled memes, we train Baseline, Image-STILT and Text-STILT on varying amounts of training memes. For each approach and at each of the training set sizes, we ran five random-restarts, resulting in 45 observations for each Baseline vs. Image-STILT and Baseline vs. Text-STILT, separately. For each random restart, we resample the training set, we define a matched pair (as required by Wilcoxon Signed-Rank test assumptions) as the performance of two models having been trained on the same set of memes. We performed a Wilcoxon Signed-Rank test across the entire range of labelled meme availability, but separately for Baseline vs. Image-STILT and Baseline vs. Text-STILT.

\section{Results}

\arrayrulecolor{lightgray}
\begin{table}[!b]
\small

\centering

\caption{Mean of Weighted F1-score, Precision and Recall and their standard deviation (in parantheses) for Baseline, Image-STILT \& Text-STILT, across 10 runs each. \label{tab_rq1}}
\begin{tabular}[]{m{0.25\columnwidth}|m{0.1\columnwidth}m{0.1\columnwidth}m{0.1\columnwidth}m{0.15\columnwidth}}
 \arrayrulecolor{black}
 \hline
 \textbf{Approach} & \textbf{F1} & \textbf{Prec} & \textbf{Rec} &\shortstack{ \textbf{p-value} \\ \tiny \textbf{vs. Baseline}} \\
 \hline
 \\[-5pt]
 \textbf{Baseline} & \shortstack{51.19 \\ \tiny (0.00393)} & \shortstack{54.86 \\ \tiny (0.0112)} & \shortstack{56.37 \\ \tiny (0.00662)} & - \\
 \textbf{Image-STILT}  & \shortstack{51.45 \\ \tiny (0.00485)} & \shortstack{ 54.96 \\ \tiny(0.0149)} & \shortstack{58.78 \\ \tiny (0.0142)} & 0.193 \\
 \textbf{Text-STILT} & \shortstack{51.78 \\ \tiny (0.00659)} & \shortstack{56.58 \\ \tiny(0.0131)} & \shortstack{57.66 \\ \tiny (0.00950)} & 0.0273 \\
\hline
\end{tabular}

\end{table}

\begin{figure}[]
  \centering
\includegraphics[width=0.85\columnwidth]{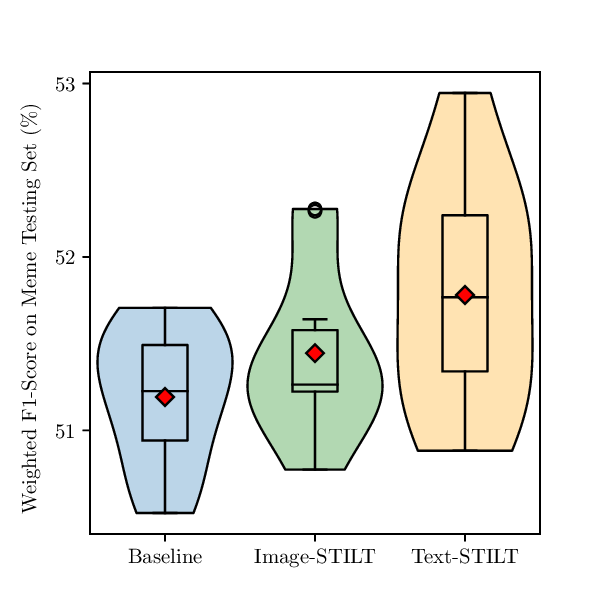}
\caption{Performance of the Baseline, Image-STILT, and Text-STILT. Box-plots indicate the 2nd - 3rd quartile range and \redmarker indicates mean performance.\label{violin_100}}
\end{figure}

\subsection{RQ1: Performance Improvement}

\begin{figure*}[]
    \centering
    \begin{subfigure}[h]{0.45\textwidth}
        \includegraphics[width=\textwidth]{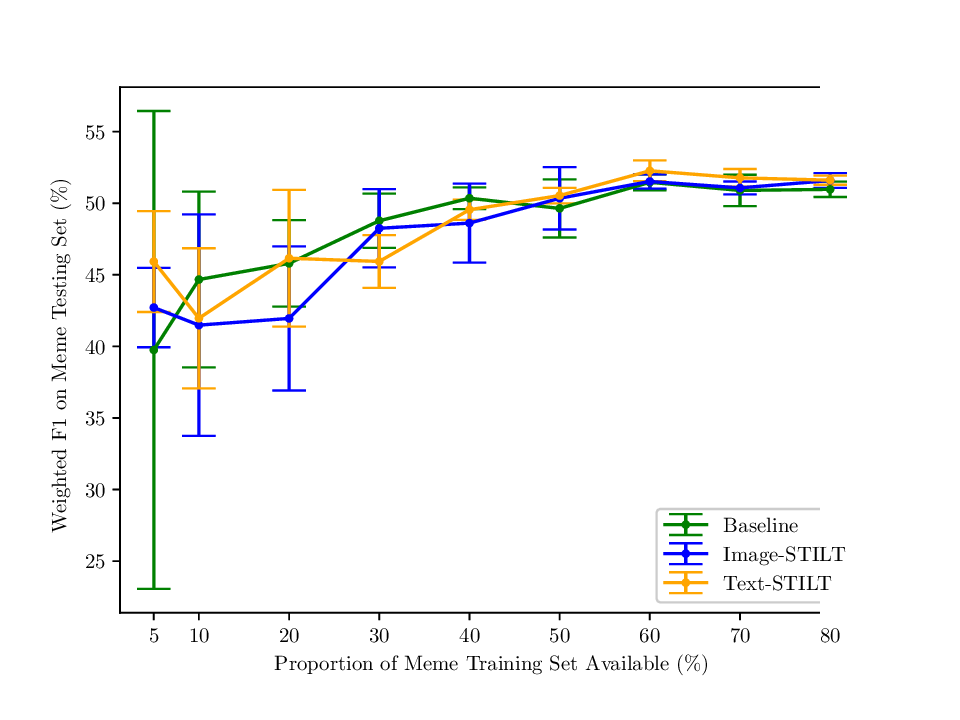}
        \caption{5\% to 80\% of Memes available.\label{fig_first_case}}
    \end{subfigure}     \begin{subfigure}[h]{0.45\textwidth}
        \includegraphics[width=\textwidth]{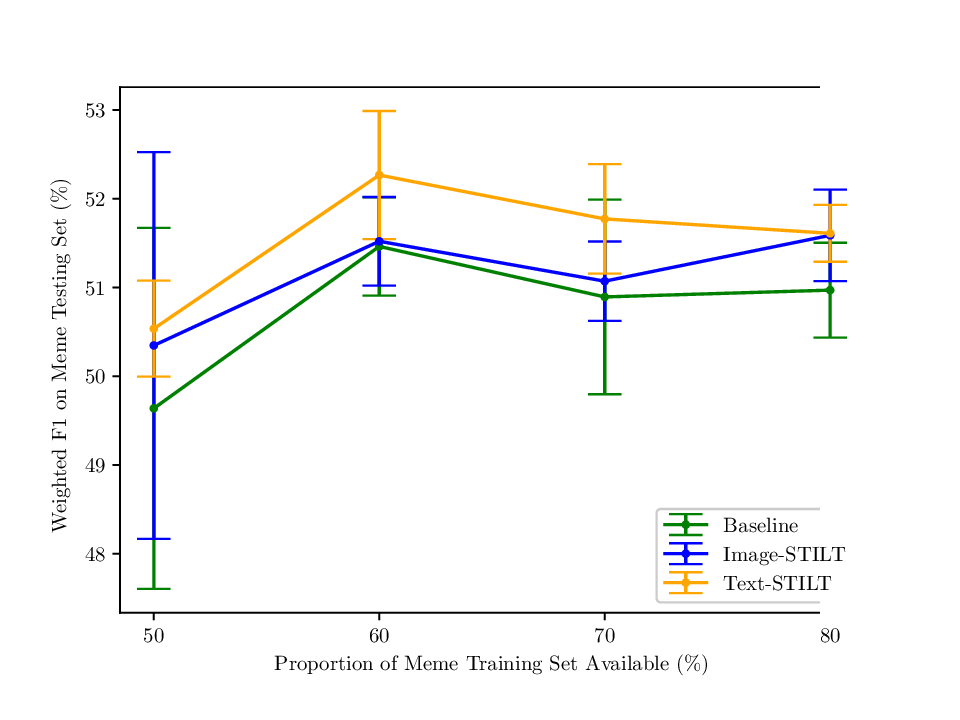}
        \caption{50\% to 80\% of Memes available.       \label{fig_second_case}}
    \end{subfigure}
    
    \caption{Baseline, Image-STILT, and Text-STILT performance across varying amount of memes available; 5 random restarts each.}
    \label{fig_sim}
\end{figure*}

\par
Text-STILT was found to outperform Baseline, at a level of statistical significance. Figure \ref{violin_100} and Table \ref{tab_rq1} show the performance distribution of each approach, with 10 random restarts each. The Wilcoxon Signed-Rank test resulted in p-values of 0.193 and 0.0273 for Baseline vs. Image-STILT and Baseline vs. Text-STILT, respectively.

To our knowledge, Text-STILT is the first approach to successfully incorporate supplementary unimodal data into the training of multimodal meme classifiers showing a statistically significant performance improvement. However, our results do not indicate why Text-STILT was effective. We posit that while each meme's semantics rely on both the image and text modalities, memes that contain longer texts and/or a textual structure that hints at the meme's overall sentiment are more accurately classified by Text-STILT (see examples in Table \ref{tab_text_vs_base}). Consider the meme in Table \ref{tab_text_vs_base}(b): While the negative component is represented visually, that is, the bottom image segment, the structure of the text ``\texttt{what people think... what it really is like...}" strongly suggests a negative inversion of something normally considered to be positive. Thus, its negative sentiment could be inferred largely from text alone. More rigorous investigation of the relationship between text and meme sentiment analysis is warranted.

Although Text-STILT significantly outperformed Baseline, Image-STILT did not. Although Image-STILT shows higher mean, maximum, and minimum performance than Baseline, the distribution (see the violin plot in Figure \ref{violin_100}) indicates that the two performed similarly. This could be attributed to the distinct role of visual symbols in memes, which derive their meaning from popular usage rather than literal connotations. Consider the memes in Table \ref{tab_vis}, each made using highly popular \textbf{meme templates}: \textit{Success Kid} \footnote{https://knowyourmeme.com/memes/success-kid-i-hate-sandcastles} and \textit{Bad Luck Brian} \footnote{https://knowyourmeme.com/memes/bad-luck-brian}, respectively. These have come to symbolise specific meanings through online usage, which is distinct from what is literally shown. In the case of \textit{Bad Luck Brian}, see Table \ref{tab_vis}(d), a teenage boy smiling in a portrait does not inherently convey tragedy, or misfortune, but this connotation stemmed from the template's usage in online discourse.

In contrast, the unimodal images in Table \ref{tab_vis} show a visual language that is less culturally specific, i.e. a serene beach has positive connotations and a disfigured \textit{zombie-esque} head conveys negative ones. The cultural specificity of visual symbols in memes likely contributed to Image-STILT's lack of significant performance improvement. These may explain similar observations by \citeauthor{suryawanshi-etal-2020-dataset} \citeyearpar{suryawanshi-etal-2020-dataset}, as discussed in Section \ref{litt_supp}, and would suggest that the transfer of visual sentiment \textit{skills} from unimodal images to multimodal meme classifiers may be inherently limited.
\newpage

\subsection{RQ2: Limited Labelled Memes}
\begin{table}[!b]
\small

\centering

\caption{Example memes which were correctly labelled by Text-STILT but not by Baseline. \label{tab_text_vs_base}}

\begin{tabularx}{\columnwidth}{l|X |X}
 \arrayrulecolor{black}
\hline

& \textbf{(a)} & \textbf{(b)} \\

\hline

  &  \includegraphics[width=\hsize]{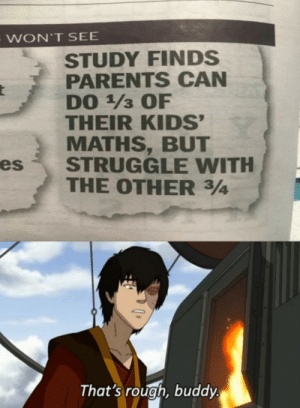} &  \includegraphics[width=\hsize]{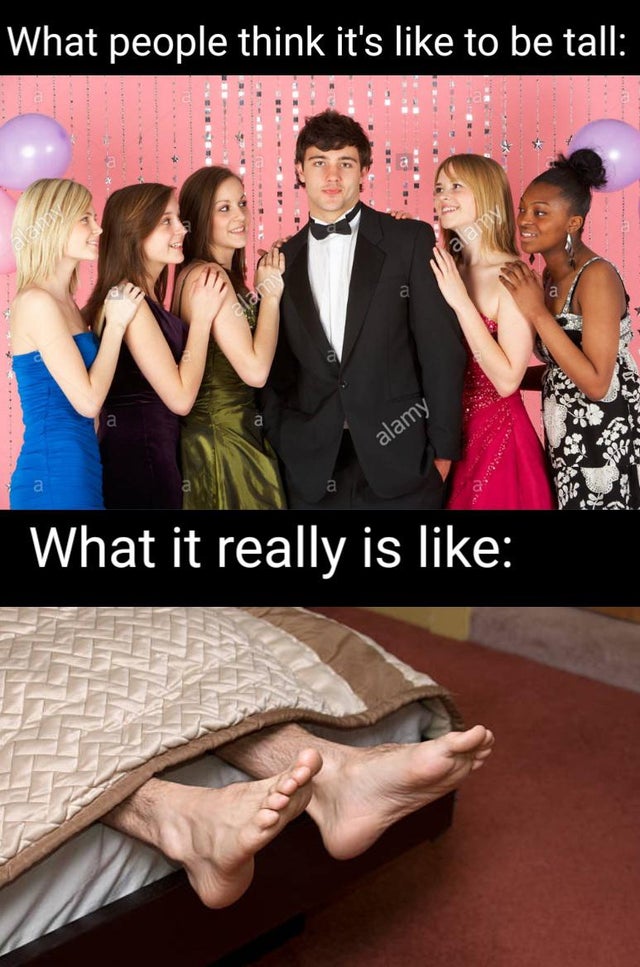} \\

\hline

\textbf{Sentiment} & Negative & Negative \\
\textbf{Predicted} & & \\

 -- \textbf{Baseline}& Positive & Neutral \\

-- \textbf{Text-STILT}  & \textbf{Negative} & \textbf{Negative}\\

\hline
\end{tabularx}
\end{table}

\begin{figure}[hbtp]
  \centering
\includegraphics[width=0.8\columnwidth]{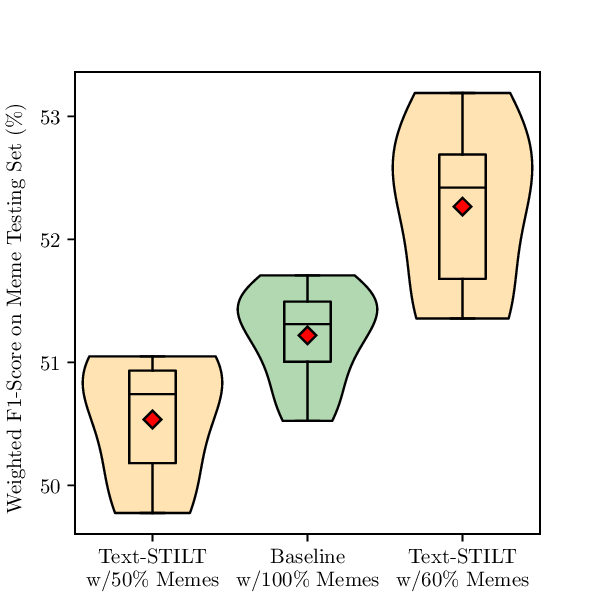}
\caption{Performance of Baseline trained on 100\% of memes available and Text-STILT trained on [50\%, 60\%] of memes available. \label{fig_violin50}}

\end{figure}

We found that Text-STILT significantly improves performance over Baseline across varying amounts of labelled meme availability between 50\% and 80\% (shown in Figure \ref{fig_second_case}). Within this range, while both intermediate training approaches consistently showed higher mean performance than Baseline, only Text-STILT showed a significant performance improvement and Image-STILT did not; p-values were 0.000109 for Baseline vs. Text-STILT and 0.0723 for Baseline vs. Image-STILT, respectively.

Based on these measurements, we found that Text-STILT was still able to outperform Baseline while using only 60\% of the available labelled memes. Figure \ref{fig_violin50} shows the performance distribution of Baseline with 100\% memes available and Text-STILT with 50\% and 60\% memes available. 

We also noted that neither Image-STILT nor Text-STILT was found to significantly improve performance over Baseline across the entire range of availability of labelled memes from 5\% to 80\%. Figure 4 shows the mean performance and standard deviation of Baseline, Image-STILT, and Text-STILT across this range. When hypothesis testing is applied across the entire range, neither Image-STILT nor Text-STILT showed statistically significant improvements over Baseline, with p-values of 0.667 and 0.985, respectively.

Although Text-STILT performed better than Baseline, the difference is small. Contemporary approaches show similar small differences in performance (see Appendix D). Furthermore, 41\% of memes in the testing set were not correctly classified by either Text-STILT and Baseline (see Appendix E). This suggests that a significant portion of memes remain a challenge to classify. This challenge might be addressed by combining Text-STILT with other supplementary training steps.

\section{Limitations and Future Works}
To generate comparable results between Baseline and Text-STILT, we kept many hyperparameters constant. Additional work would be required to determine the maximum achievable performance of Text-STILT on the chosen task.

Despite the efficacy of Text-STILT over Image-STILT, these results do not suggest that only the text modality is significant in classifying multimodal memes. Previous works have performed modality ablation studies in this problem space \cite{blue,momenta,keswani-etal-2020-iitk-semeval} with multimodal architectures remaining the apparent state of the practise. All models in this work are similarly multimodal. In the future, we plan to reformulate Image-STILT with respect to the approach and data used to isolate the cause of its non-performance on the downstream task.  Furthermore, we did not test Text-STILT on classifiers that represent the image modality of a meme in textual forms, as others did \cite{singh-etal-2022-combining,momenta}.

Notwithstanding our results, Text-STILT may not benefit all multimodal meme classifiers. \citet{phang2019sentence} showed that STILT offers varying degrees of benefit depending on the encoders chosen. Future work is needed to verify if these observations hold across the wide range of pretrained encoders commonly used in meme classifiers. In particular, some modifications to unimodal STILTs are needed to be applied to single-stream multimodal encoders, as those used in other works.

Furthermore, \citet{pruksachatkun-etal-2020-intermediate}  showed that intermediate training benefits various text-only tasks differently. We have yet to identify other meme classification tasks that would benefit from unimodal STILTs. Thus, we plan to conduct more extensive experimentation to validate the effectiveness of Text-STILT on other meme classification tasks, e.g. pairing hate-speech detection in text \cite{toraman-etal-2022-large} as an intermediate task for hateful meme detection \cite{hateful_data}.

\section{Conclusion}
In this work, we addressed the challenge of training multimodal meme sentiment classifiers on a limited number of labelled memes by incorporating unimodal sentiment analysis data. We did so by proposing the first instance of STILT that applies unimodal intermediate tasks to a multimodal target task. Specifically, we tested image-only and text-only sentiment classification as intermediate tasks in training a meme sentiment classifier. We showed that this approach worked -- unimodal text improved meme classification performance to a statistically significant degree. This novel approach allowed us to train a meme classifier that outperforms meme-only finetuning with only 60\% as many labelled meme samples. As possible explanations for our observations, we discuss apparent similarities and differences in the roles of image and text modalities between unimodal and multimodal sentiment analysis tasks.

\clearpage

\section*{Acknowledgments}
This work was conducted with the financial support of the Science Foundation Ireland Centre for Research Training in Digitally-Enhanced Reality (d-real) under Grant No. 18/CRT/6224; and the provision of computational facilities and support from the Irish Centre for High-End Computing (ICHEC).

\bibliographystyle{acl_natbib}

\clearpage

\appendix

\section{Hyperparameters and Settings}
\arrayrulecolor{lightgray}
\begin{table}[!h]
\vspace{20pt}
    \centering

    \begin{tabularx}{\columnwidth}{m{1.in}|>{\centering\arraybackslash}X |>{\centering\arraybackslash}X}

    \arrayrulecolor{black}
    \hline
    \textbf{} & \multicolumn{2}{c}{\textbf{Input}} \\
    &\textbf{Memes} & \textbf{Unimodal}\\
    \hline
    \arrayrulecolor{lightgray}
    LR Scheduling & \multicolumn{2}{c}{Cosine Annealing}\\
    Loss & \multicolumn{2}{c}{Negative Log-Likelihood} \\
    Learning Rate & $1.5\mathrm{e}{-5}$ to $5\mathrm{e}{-5}$ &$5\mathrm{e}{-6}$ to $1\mathrm{e}{-5}$\\
    Max Epochs & 40 & 60 \\
    Optimizer & \multicolumn{2}{c}{AdamW}\\
    Betas & \multicolumn{2}{c}{[0.5 , 0.9]} \\
    Weight Decay & \multicolumn{2}{c}{0.9} \\
    AMSGrad &\multicolumn{2}{c}{False} \\
    Dropout Rate & \multicolumn{2}{c}{0.2} \\
    Early-Stopping \tiny (per Meme Validation set) & Min Loss & Max Wei. F1\\
    \arrayrulecolor{black} \hline

    \end{tabularx}
    \caption{Hyperparameter values and settings used during model training by input type.}
    \label{tab:my_label}
\end{table}

\section{Metric: Weighted F1-Score}
The performance of our models are measured by Weighted F1-Score, inline with the reporting set by the authors of the Memotion 2.0 dataset \cite{memo2_report}. The F1-Score is the harmonic mean of precision and recall, equally representing both. ``Weighted'' here denotes that the F1-score is first computed per-class and then averaged while weighted by the proportion of occurrences of each class in the ground truth labels. We compute this using PyTorch's implementation \texttt{multiclass\_f1\_score}. Class-wise F1-scores, $F1_c$ where $c \in [1,2,3]$, are computed as:

\begin{equation}
\begin{gathered}
precision_c = \frac{TP_c}{TP_c + FP_c}
\\
recall_c = \frac{TP_c}{TP_c + FN_c}
\\
F1_c = 2 \times \frac{ (precision_c \times recall_c)}{ (precision_c + recall_c)}
\end{gathered}
\end{equation}

Where $TP_c, FP_c, FN_c$ are the count of true positives, false positives and false negatives, respectively. The Weighted F1-score is computed as the weighted average of $F1_c$:
\begin{equation}
\begin{gathered}
w_c = \frac{N_c}{\sum_{c=0}^{C}N_c}
\\
F1 = \frac{\sum_{c=0}^{C} w_c F1_c}{C}
\end{gathered}
\end{equation}
Where $N_c$ is the number of samples with the ground truth label $c$ in the testing set.
The Weighted F1 is often used when the classes are imbalanced -- the training, validation and testing sets of Memotion 2.0 show significant and varying class imblance -- as it takes into account the relative importance of each class. Note that this weighted averaging could result in an F1-score that is not between the Precision and Recall scores.

\section{Architectural Details}
Our models are based on the Baseline model proposed by \citet{hazman_aics} and we similarly utilise the Image and Text Encoders from the pretrained ViT--B/16 CLIP model to generate representations of each modality.

\begin{equation}\label{global_encoding}
\begin{gathered}
    F_I = ImageEncoder(Image)
    \\
    F_T = TextEncoder(Text) 
\end{gathered}
\end{equation}
Where each $F_I$ and $F_T$ is a 512-digit embedding of the image and text modalities, respectively, from CLIP's embedding space that aligns images with their corresponding text captions \cite{clip}.

For unimodal inputs, the encoder for the missing modality is fed a blank input, i.e. when finetuning on unimodal images, the text input is defined as a string containing no characters i.e. ``":
\begin{equation}\label{global_encoding}
\begin{gathered}
    F_I = ImageEncoder(Image)
    \\
    F_T = TextEncoder(``") 
\end{gathered}
\end{equation}
Conversely, when finetuning on unimodal texts, the image input is defined as a $3\times224\times224$ matrix of zeros, or equivalently, JPEG file with all pixels set to black.
\begin{equation}\label{global_encoding}
\begin{gathered}
    F_I = ImageEncoder(O_{3\times224\times224})
    \\
    F_T = TextEncoder(Text) 
\end{gathered}
\end{equation}

 For each modality, we added dropout and normalisation:
\begin{equation}
\begin{gathered}
    f_I = Norm(Dropout(F_I))
    \\
    f_T = Norm(Dropout(F_T))
    \end{gathered}
\end{equation}

where $Norm()$ is PyTorch's BatchNorm1D and $Dropout$ has a rate of 0.2. These modality representations $f_I$ and $f_T$  are then placed into an attentive fusion mechanism proposed by \citeauthor{gu_fusion} \citeyearpar{gu_fusion} and used by \citeauthor{mhameme} \citeyearpar{mhameme,momenta} and \citeauthor{hazman_aics} \citeyearpar{hazman_aics}. The embedding representation for each modality is passed through four dense layers of reducing sizes $[256, 64, 8, 1]$, $Dense_{i}$ and $Dense_{t}$ for the image and text modalities, respectively. Then, softmax is applied on the output of each stack is to generate a weighted score for each modality. Per \cite{gu_fusion}:
\begin{equation}\label{ATMF}
\begin{gathered}
D_i = Dense_i(f_I)\\
D_t = Dense_t(f_T)\\
[s_i,s_t] = softmax(W_f[D_i, D_t] + b_f)\\
S_i = (1 + s_i)\\
S_t = (1 + s_t)\\
F_{MM} = tanh(W_r[S_i(f_I) ,S_t(f_T)] + b_r)
\end{gathered}
\end{equation}

We added a dropout and normalisation step onto the fused multimodal representation, $F_{MM}$:
\begin{equation}
    f_{MM} = Norm(Dropout(F_{MM}))
\end{equation}

The predicted logits of each class is given by passing $f_{MM}$ a dense network of GeLU-activated layers with sizes [1024, 256,3]:
\begin{equation}
\begin{gathered}
    X_{MM} = tanh(W_{mm}(W_x(f_{MM}) + b_x) + b_{mm})
    \\
    logits = W_{l}(X_{MM}) + b_l
    \end{gathered}
\end{equation}

The model is fitted by minimising the mean multiclass Cross Entropy Loss per PyTorch's definition:
\begin{equation}
\begin{gathered}
    l_n = - w_{y_n} \log \frac{exp(x_{n,y_n})}{\sum_{i=1}^{C} exp(x_{n,c})} \dot y_n
    \\
    L = \sum_{n=1}^{N} \frac{1}{\sum_{n=1}^{N} w_{y_n}} l_n
    \end{gathered}
\end{equation}
Where $x_{n,y_n}$ is the logits for each class and $y_n$ is the target label of a given sample $n$ of total $N$ samples in a minibatch; c is the class in [Negative, Neutral and Positive] and C is the number of classes. The loss for each sample is weighted by:
\begin{equation}
\begin{gathered}
    w_{y_n} = 1 - \frac{N_{y_n}}{\sum_{{y_n}=0}^{C} N_{y_n}}
    \end{gathered}
\end{equation}

Where $N_0,N_1,N_2$ are the number of training samples labelled with Negative, Neutral and Positive sentiment, respectively.

\section{Performance Benchmarking}
Current competing approaches show a small spread of Weigthed F1-scores (see Table \ref{tab_benchmark}) and the performance improvement offered by Text-STILT is similarly small. This small range of performances in contemporary approaches suggests that there is still a significant portion of memes that remain a challenge to classify.

\begin{table}[!t]
\small
\centering

\caption{The mean and maximum Weighted F1-scores from our Baseline and Text-STILT approaches against various SOTA solutions. \label{tab_benchmark}}
\begin{tabular}{l|c}
 
 \arrayrulecolor{black}
 \hline
 \textbf{Solution} & \textbf{Weighted F1 (\%)} \\
 \hline
 
\citet{blue} & \textbf{53.18} \\
\textit{Text-STILT w/ 60\% (Max)} & 53.15\\
\citet{Duan2022BROWALLIAAM} & 52.55 \\
\textit{Text-STILT w/ 60\% (Mean)} & 52.45\\
\textit{Our Baseline (Max)} & 51.70\\
\textit{Our Baseline (Mean)} & 51.19\\
\citet{Zhuang2022YetAM} & 50.88 \\
\citet{Phan2022LittleFA} & 50.81 \\
Greeny (via \citeauthor{memo2_report}, \citeyear{memo2_report}) & 50.37 \\
\citet{hazman_aics} & 50.35 \\
\citet{memo2_amazon} & 50.25 \\
\citet{NGUYEN2022HCILabAM} & 49.95\\
\hline
\end{tabular}

\end{table}

\section{Contingency Table: Baseline vs. Text-STILT}

\begin{table}[!h]
\caption{Contingency Table between similarly performing Text-STILT (trained with 60\% memes) and Baseline (trained with 100\% memes).} \label{tab_cont}
\centering
\begin{tabularx}{\columnwidth}{>{\centering}X >{\centering}X |>{\centering}X X}
\arrayrulecolor{black}
\hline
 & &  \multicolumn{2}{c}{ \centering \textbf{Baseline}}  \\
  & & \centering \textbf{Correct} & \textbf{Wrong}  \\
  
\hline
\multirow{2}{*}{\shortstack{\textbf{Text-} \\ \textbf{STILT}}} &  \textbf{Correct} & \centering 610 & 146 \\
 & \textbf{Wrong}& \centering 136 & 608\\
 \hline
\end{tabularx}
\end{table}

Table \ref{tab_cont} shows the contingency table -- as one would prepare for a McNemar's Test between two classifiers \cite{McNemar1947} -- between the model trained with Text-STILT on 60\% Memes and Baseline trained on 100\% Memes available which had the most similar performance. While the two models performed similarly in terms of Weighted F1-scores, Text-STILT correctly classified a notable number of memes that Baseline did not and vice versa. Examples of such memes are discussed in Section 4.1. Furthermore, approximately 40\% of memes in the testing set were incorrectly classified by both models. This suggests that these memes convey sentiment in a way that cannot be reliably predicted by either approach.

\end{document}

%% file: figures/diag_train_wide.tex
\tikzset{every picture/.style={line width=0.75pt}} 

\begin{tikzpicture}[x=0.75pt,y=0.75pt,yscale=-1,xscale=1]

\draw  [fill={rgb, 255:red, 178; green, 216; blue, 178 }  ,fill opacity=1 ] (456.33,462.17) -- (613.2,462.17) -- (649,512.17) -- (613.2,562.17) -- (456.33,562.17) -- cycle ;
\draw  [fill={rgb, 255:red, 178; green, 216; blue, 178 }  ,fill opacity=1 ] (454,329.4) -- (612.47,329.4) -- (651.33,374.2) -- (612.47,419) -- (454,419) -- cycle ;
\draw  [fill={rgb, 255:red, 255; green, 227; blue, 178 }  ,fill opacity=1 ] (452,231.37) -- (602.07,231.37) -- (641.27,253.72) -- (602.07,276.06) -- (452,276.06) -- cycle ;
\draw  [fill={rgb, 255:red, 178; green, 216; blue, 178 }  ,fill opacity=1 ] (361.67,94.53) -- (612.66,94.53) -- (652.07,139.73) -- (612.66,184.93) -- (361.67,184.93) -- cycle ;
\draw  [fill={rgb, 255:red, 187; green, 213; blue, 232 }  ,fill opacity=1 ] (51.47,89.1) -- (321.05,89.1) -- (361.67,111.81) -- (321.05,134.53) -- (51.47,134.53) -- cycle ;
\draw  [fill={rgb, 255:red, 255; green, 227; blue, 178 }  ,fill opacity=1 ] (51.47,144.77) -- (319.47,144.77) -- (360.67,166.85) -- (319.47,188.93) -- (51.47,188.93) -- cycle ;
\draw  [fill={rgb, 255:red, 255; green, 227; blue, 178 }  ,fill opacity=1 ] (252.13,231.37) -- (411.6,231.37) -- (451.6,253.72) -- (411.6,276.06) -- (252.13,276.06) -- cycle ;
\draw  [fill={rgb, 255:red, 255; green, 227; blue, 178 }  ,fill opacity=1 ] (51.47,231.37) -- (210.43,231.37) -- (250.27,253.72) -- (210.43,276.06) -- (51.47,276.06) -- cycle ;
\draw  [fill={rgb, 255:red, 187; green, 213; blue, 232 }  ,fill opacity=1 ] (51.47,323.77) -- (209.32,323.77) -- (250.27,345.78) -- (209.32,367.8) -- (51.47,367.8) -- cycle ;
\draw  [fill={rgb, 255:red, 255; green, 227; blue, 178 }  ,fill opacity=1 ] (51.47,377.98) -- (409.6,377.98) -- (450.4,400.54) -- (409.6,423.1) -- (51.47,423.1) -- cycle ;
\draw  [fill={rgb, 255:red, 187; green, 213; blue, 232 }  ,fill opacity=1 ] (254.47,323.77) -- (412.32,323.77) -- (453.27,345.78) -- (412.32,367.8) -- (254.47,367.8) -- cycle ;
\draw  [fill={rgb, 255:red, 255; green, 227; blue, 178 }  ,fill opacity=1 ] (51.47,517.2) -- (209.32,517.2) -- (250.27,539.21) -- (209.32,561.23) -- (51.47,561.23) -- cycle ;
\draw  [fill={rgb, 255:red, 255; green, 227; blue, 178 }  ,fill opacity=1 ] (254.47,517.2) -- (412.32,517.2) -- (453.27,539.21) -- (412.32,561.23) -- (254.47,561.23) -- cycle ;
\draw  [fill={rgb, 255:red, 187; green, 213; blue, 232 }  ,fill opacity=1 ] (51.47,464.38) -- (409.6,464.38) -- (450.4,486.94) -- (409.6,509.5) -- (51.47,509.5) -- cycle ;
\draw  [color={rgb, 255:red, 155; green, 155; blue, 155 }  ,draw opacity=1 ] (668,88.58) .. controls (668,85.4) and (670.57,82.83) .. (673.74,82.83) -- (769.92,82.83) .. controls (773.1,82.83) and (775.67,85.4) .. (775.67,88.58) -- (775.67,212.09) .. controls (775.67,215.26) and (773.1,217.83) .. (769.92,217.83) -- (673.74,217.83) .. controls (670.57,217.83) and (668,215.26) .. (668,212.09) -- cycle ;
\draw  [fill={rgb, 255:red, 187; green, 213; blue, 232 }  ,fill opacity=1 ][line width=1.5]  (675,118.12) -- (756.7,118.12) -- (769.01,129.76) -- (756.7,141.4) -- (675,141.4) -- cycle ;
\draw  [fill={rgb, 255:red, 255; green, 227; blue, 178 }  ,fill opacity=1 ][line width=1.5]  (675,148.96) -- (756.7,148.96) -- (769.01,160.6) -- (756.7,172.24) -- (675,172.24) -- cycle ;
\draw  [fill={rgb, 255:red, 178; green, 216; blue, 178 }  ,fill opacity=1 ][line width=1.5]  (675,183.3) -- (759.16,183.3) -- (771.83,194.94) -- (759.16,206.57) -- (675,206.57) -- cycle ;
\draw  [color={rgb, 255:red, 155; green, 155; blue, 155 }  ,draw opacity=1 ] (45.83,90.11) .. controls (45.83,86.82) and (48.5,84.16) .. (51.79,84.16) -- (648.88,84.16) .. controls (652.17,84.16) and (654.83,86.82) .. (654.83,90.11) -- (654.83,189.88) .. controls (654.83,193.17) and (652.17,195.83) .. (648.88,195.83) -- (51.79,195.83) .. controls (48.5,195.83) and (45.83,193.17) .. (45.83,189.88) -- cycle ;
\draw  [color={rgb, 255:red, 155; green, 155; blue, 155 }  ,draw opacity=1 ] (45.83,324.11) .. controls (45.83,320.82) and (48.5,318.16) .. (51.79,318.16) -- (648.88,318.16) .. controls (652.17,318.16) and (654.83,320.82) .. (654.83,324.11) -- (654.83,423.88) .. controls (654.83,427.17) and (652.17,429.83) .. (648.88,429.83) -- (51.79,429.83) .. controls (48.5,429.83) and (45.83,427.17) .. (45.83,423.88) -- cycle ;
\draw  [color={rgb, 255:red, 155; green, 155; blue, 155 }  ,draw opacity=1 ] (46.5,464.11) .. controls (46.5,460.82) and (49.17,458.16) .. (52.46,458.16) -- (649.54,458.16) .. controls (652.83,458.16) and (655.5,460.82) .. (655.5,464.11) -- (655.5,563.88) .. controls (655.5,567.17) and (652.83,569.83) .. (649.54,569.83) -- (52.46,569.83) .. controls (49.17,569.83) and (46.5,567.17) .. (46.5,563.88) -- cycle ;
\draw  [color={rgb, 255:red, 155; green, 155; blue, 155 }  ,draw opacity=1 ] (46.47,228.59) .. controls (46.47,226.84) and (47.89,225.41) .. (49.64,225.41) -- (652.29,225.41) .. controls (654.04,225.41) and (655.47,226.84) .. (655.47,228.59) -- (655.47,281.82) .. controls (655.47,283.58) and (654.04,285) .. (652.29,285) -- (49.64,285) .. controls (47.89,285) and (46.47,283.58) .. (46.47,281.82) -- cycle ;

\draw (101.37,158.43) node [anchor=north west][inner sep=0.75pt]  [font=\large] [align=left] {Text Encoder Pretraining};
\draw (90.37,102.6) node [anchor=north west][inner sep=0.75pt]  [font=\large] [align=left] {Image Encoder Pretraining};
\draw (422.27,118.93) node [anchor=north west][inner sep=0.75pt]  [font=\large] [align=left] {\begin{minipage}[lt]{98.64pt}\setlength\topsep{0pt}
\begin{center}
Meme Sentiment \\Classification
\end{center}

\end{minipage}};
\draw (86.17,235.22) node [anchor=north west][inner sep=0.75pt]  [font=\large] [align=left] {\begin{minipage}[lt]{77.56pt}\setlength\topsep{0pt}
\begin{center}
Text Encoder \\Pretraining
\end{center}

\end{minipage}};
\draw (488.53,235.22) node [anchor=north west][inner sep=0.75pt]  [font=\large] [align=left] {\begin{minipage}[lt]{68.04pt}\setlength\topsep{0pt}
Target Task
\begin{center}
(Finetuning)
\end{center}

\end{minipage}};
\draw (258.13,234.37) node [anchor=north west][inner sep=0.75pt]  [font=\large] [align=left] {\begin{minipage}[lt]{125.19pt}\setlength\topsep{0pt}
\begin{center}
{\small Supervised Intermediate }\\{\small Task ($\neq$ Target Task)}
\end{center}

\end{minipage}};
\draw (476.57,355.7) node [anchor=north west][inner sep=0.75pt]  [font=\large] [align=left] {\begin{minipage}[lt]{95.24pt}\setlength\topsep{0pt}
\begin{center}
Meme Sentiment\\Classification
\end{center}

\end{minipage}};
\draw (279.37,327.28) node [anchor=north west][inner sep=0.75pt]  [font=\large] [align=left] {\begin{minipage}[lt]{95.25pt}\setlength\topsep{0pt}
\begin{center}
Image Sentiment\\Classification
\end{center}

\end{minipage}};
\draw (82,327.28) node [anchor=north west][inner sep=0.75pt]  [font=\large] [align=left] {\begin{minipage}[lt]{85.73pt}\setlength\topsep{0pt}
\begin{center}
Image Encoder\\Pretraining
\end{center}

\end{minipage}};
\draw (201.17,382.17) node [anchor=north west][inner sep=0.75pt]  [font=\large] [align=left] {\begin{minipage}[lt]{74.16pt}\setlength\topsep{0pt}
\begin{center}
Text Encoder\\Pretraining
\end{center}

\end{minipage}};
\draw (480.57,493.67) node [anchor=north west][inner sep=0.75pt]  [font=\large] [align=left] {\begin{minipage}[lt]{95.24pt}\setlength\topsep{0pt}
\begin{center}
Meme Sentiment\\Classification
\end{center}

\end{minipage}};
\draw (278.37,520.71) node [anchor=north west][inner sep=0.75pt]  [font=\large] [align=left] {\begin{minipage}[lt]{87.08pt}\setlength\topsep{0pt}
\begin{center}
Text Sentiment \\Classification
\end{center}

\end{minipage}};
\draw (192.2,468.63) node [anchor=north west][inner sep=0.75pt]  [font=\large] [align=left] {\begin{minipage}[lt]{85.73pt}\setlength\topsep{0pt}
\begin{center}
Image Encoder\\Pretraining
\end{center}

\end{minipage}};
\draw (92.2,520.71) node [anchor=north west][inner sep=0.75pt]  [font=\large] [align=left] {\begin{minipage}[lt]{74.16pt}\setlength\topsep{0pt}
\begin{center}
Text Encoder\\Pretraining
\end{center}

\end{minipage}};
\draw (54.47,43) node [anchor=north west][inner sep=0.75pt]   [align=left] {\textbf{{\Large Baseline}}\\\textbf{(Meme Classifier Finetuning)}};
\draw (52.47,206.47) node [anchor=north west][inner sep=0.75pt]   [align=left] {\textbf{{\Large \citeauthor{phang2019sentence}'s \citeyearpar{phang2019sentence} STILT}}};
\draw (52.47,297.47) node [anchor=north west][inner sep=0.75pt]   [align=left] {\textbf{{\Large Image-STILT}}};
\draw (52.47,442.27) node [anchor=north west][inner sep=0.75pt]   [align=left] {\textbf{{\Large Text-STILT}}};
\draw (670,91.58) node [anchor=north west][inner sep=0.75pt]  [font=\normalsize] [align=left] {\textbf{Task Modality}};
\draw (676,121.76) node [anchor=north west][inner sep=0.75pt]  [font=\normalsize] [align=left] {\textbf{Visual}};
\draw (676,152.6) node [anchor=north west][inner sep=0.75pt]  [font=\normalsize] [align=left] {\textbf{Textual}};
\draw (676,186.94) node [anchor=north west][inner sep=0.75pt]  [font=\normalsize] [align=left] {\textbf{Multimodal}};

\end{tikzpicture}

%% file: figures/diag_arch.tex
\tikzset{every picture/.style={line width=0.75pt}} 

\definecolor{yel}{HTML}{FFE3B2}
\definecolor{blu}{HTML}{B2D8B2}
\definecolor{grn}{HTML}{BBD5E8}

\tikzset{every picture/.style={line width=0.75pt}} 

\begin{tikzpicture}[x=0.75pt,y=0.75pt,yscale=-1,xscale=1,trim left={(2.8,0)}]
\draw [fill=grn]  (373.67,104.67) -- (378.33,104.67) -- (378.33,148.83) -- (373.67,148.83) -- cycle ;
\draw [fill=grn]  (383.33,110.33) -- (387.33,110.33) -- (387.33,141.5) -- (383.33,141.5) -- cycle ;
\draw [fill=grn]  (392.67,114.67) -- (395.67,114.67) -- (395.67,135.17) -- (392.67,135.17) -- cycle ;
\draw  [fill=grn] (400,120.33) -- (403,120.33) -- (403,129.83) -- (400,129.83) -- cycle ;
\draw  [fill=yel] (375.33,159) -- (380,159) -- (380,203.17) -- (375.33,203.17) -- cycle ;
\draw  [fill=yel] (385,164.67) -- (389,164.67) -- (389,195.83) -- (385,195.83) -- cycle ;
\draw  [fill=yel] (394.33,169) -- (397.33,169) -- (397.33,189.5) -- (394.33,189.5) -- cycle ;
\draw [fill=yel]  (401.67,174.67) -- (404.67,174.67) -- (404.67,184.17) -- (401.67,184.17) -- cycle ;

\draw   (411.33,223.75) .. controls (411.33,219.29) and (415.06,215.67) .. (419.67,215.67) .. controls (424.23,215.67) and (428,219.29) .. (428,223.75) .. controls (428,228.21) and (424.23,231.83) .. (419.67,231.83) .. controls (415.06,231.83) and (411.33,228.21) .. (411.33,223.75) -- cycle ; \draw   (413.77,218.03) -- (425.56,229.47) ; \draw   (425.56,218.03) -- (413.77,229.47) ;
\draw   (410,83.65) .. controls (410,79.19) and (413.73,75.57) .. (418.33,75.57) .. controls (422.94,75.57) and (426.67,79.19) .. (426.67,83.65) .. controls (426.67,88.12) and (422.94,91.73) .. (418.33,91.73) .. controls (413.73,91.73) and (410,88.12) .. (410,83.65) -- cycle ; \draw   (412.44,77.94) -- (424.23,89.37) ; \draw   (424.23,77.94) -- (412.44,89.37) ;
\draw    (334,193.5) -- (372.67,193.5) ;
\draw [shift={(372.67,193.5)}, rotate = 180] [color={rgb, 255:red, 0; green, 0; blue, 0 }  ][line width=0.75]    (10.93,-3.29) .. controls (6.95,-1.4) and (3.31,-0.3) .. (0,0) .. controls (3.31,0.3) and (6.95,1.4) .. (10.93,3.29)   ;
\draw    (405.33,178.17) -- (419,178.5) -- (419,215) ;
\draw [shift={(419,213)}, rotate = 270] [color={rgb, 255:red, 0; green, 0; blue, 0 }  ][line width=0.75]    (10.93,-3.29) .. controls (6.95,-1.4) and (3.31,-0.3) .. (0,0) .. controls (3.31,0.3) and (6.95,1.4) .. (10.93,3.29)   ;
\draw    (403.33,126) -- (419,126) -- (419,92) ;
\draw [shift={(419,95)}, rotate = 90] [color={rgb, 255:red, 0; green, 0; blue, 0 }  ][line width=0.75]    (10.93,-3.29) .. controls (6.95,-1.4) and (3.31,-0.3) .. (0,0) .. controls (3.31,0.3) and (6.95,1.4) .. (10.93,3.29)   ;
\draw    (333.33,224.5) -- (409.67,224.5) ;
\draw [shift={(407.67,224.5)}, rotate = 180] [color={rgb, 255:red, 0; green, 0; blue, 0 }  ][line width=0.75]    (10.93,-3.29) .. controls (6.95,-1.4) and (3.31,-0.3) .. (0,0) .. controls (3.31,0.3) and (6.95,1.4) .. (10.93,3.29)   ;
\draw    (333.67,83.65) -- (407,83.65) ;
\draw [shift={(407,83.65)}, rotate = 180] [color={rgb, 255:red, 0; green, 0; blue, 0 }  ][line width=0.75]    (10.93,-3.29) .. controls (6.95,-1.4) and (3.31,-0.3) .. (0,0) .. controls (3.31,0.3) and (6.95,1.4) .. (10.93,3.29)   ;
\draw    (333.09,113) -- (371,113) ;
\draw [shift={(370,113)}, rotate = 180] [color={rgb, 255:red, 0; green, 0; blue, 0 }  ][line width=0.75]    (10.93,-3.29) .. controls (6.95,-1.4) and (3.31,-0.3) .. (0,0) .. controls (3.31,0.3) and (6.95,1.4) .. (10.93,3.29)   ;

\draw  [fill=grn] (325.67,65.09) .. controls (325.67,64.33) and (326.29,63.71) .. (327.05,63.71) -- (331.21,63.71) .. controls (331.97,63.71) and (332.6,64.33) .. (332.6,65.09) -- (332.6,121.45) .. controls (332.6,122.21) and (331.97,122.83) .. (331.21,122.83) -- (327.05,122.83) .. controls (326.29,122.83) and (325.67,122.21) .. (325.67,121.45) -- cycle ;
\draw  [fill=white] (327.57,68.17) .. controls (327.57,67.21) and (328.34,66.43) .. (329.3,66.43) .. controls (330.26,66.43) and (331.03,67.21) .. (331.03,68.17) .. controls (331.03,69.12) and (330.26,69.9) .. (329.3,69.9) .. controls (328.34,69.9) and (327.57,69.12) .. (327.57,68.17) -- cycle ;
\draw  [fill=white] (327.7,73.9) .. controls (327.7,72.94) and (328.48,72.17) .. (329.43,72.17) .. controls (330.39,72.17) and (331.17,72.94) .. (331.17,73.9) .. controls (331.17,74.86) and (330.39,75.63) .. (329.43,75.63) .. controls (328.48,75.63) and (327.7,74.86) .. (327.7,73.9) -- cycle ;
\draw  [fill=white] (327.7,79.37) .. controls (327.7,78.41) and (328.48,77.63) .. (329.43,77.63) .. controls (330.39,77.63) and (331.17,78.41) .. (331.17,79.37) .. controls (331.17,80.32) and (330.39,81.1) .. (329.43,81.1) .. controls (328.48,81.1) and (327.7,80.32) .. (327.7,79.37) -- cycle ;
\draw [fill=white]  (328.58,101.67) .. controls (328.58,101.35) and (328.84,101.1) .. (329.15,101.1) .. controls (329.46,101.1) and (329.72,101.35) .. (329.72,101.67) .. controls (329.72,101.98) and (329.46,102.23) .. (329.15,102.23) .. controls (328.84,102.23) and (328.58,101.98) .. (328.58,101.67) -- cycle ;
\draw  [fill=white] (328.57,104.88) .. controls (328.57,104.57) and (328.82,104.32) .. (329.13,104.32) .. controls (329.45,104.32) and (329.7,104.57) .. (329.7,104.88) .. controls (329.7,105.2) and (329.45,105.45) .. (329.13,105.45) .. controls (328.82,105.45) and (328.57,105.2) .. (328.57,104.88) -- cycle ;
\draw  [fill=white] (328.56,98.27) .. controls (328.56,97.96) and (328.82,97.71) .. (329.13,97.71) .. controls (329.44,97.71) and (329.7,97.96) .. (329.7,98.27) .. controls (329.7,98.59) and (329.44,98.84) .. (329.13,98.84) .. controls (328.82,98.84) and (328.56,98.59) .. (328.56,98.27) -- cycle ;
\draw  [fill=white] (327.57,117.37) .. controls (327.57,116.41) and (328.34,115.63) .. (329.3,115.63) .. controls (330.26,115.63) and (331.03,116.41) .. (331.03,117.37) .. controls (331.03,118.32) and (330.26,119.1) .. (329.3,119.1) .. controls (328.34,119.1) and (327.57,118.32) .. (327.57,117.37) -- cycle ;
\draw [fill=white]  (327.7,111.37) .. controls (327.7,110.41) and (328.48,109.63) .. (329.43,109.63) .. controls (330.39,109.63) and (331.17,110.41) .. (331.17,111.37) .. controls (331.17,112.32) and (330.39,113.1) .. (329.43,113.1) .. controls (328.48,113.1) and (327.7,112.32) .. (327.7,111.37) -- cycle ;
\draw  [fill=white] (327.57,85.62) .. controls (327.57,84.66) and (328.34,83.89) .. (329.3,83.89) .. controls (330.26,83.89) and (331.03,84.66) .. (331.03,85.62) .. controls (331.03,86.58) and (330.26,87.35) .. (329.3,87.35) .. controls (328.34,87.35) and (327.57,86.58) .. (327.57,85.62) -- cycle ;
\draw  [fill=white] (327.7,91.35) .. controls (327.7,90.4) and (328.48,89.62) .. (329.43,89.62) .. controls (330.39,89.62) and (331.17,90.4) .. (331.17,91.35) .. controls (331.17,92.31) and (330.39,93.09) .. (329.43,93.09) .. controls (328.48,93.09) and (327.7,92.31) .. (327.7,91.35) -- cycle ;

\draw   [fill=yel] (326,186) .. controls (326,186) and (326.62,185.33) .. (327.39,185.33) -- (331.54,185.33) .. controls (332.31,185.33) and (332.93,185.95) .. (332.93,186.72) -- (332.93,243.07) .. controls (332.93,243.84) and (332.31,244.46) .. (331.54,244.46) -- (327.39,244.46) .. controls (326.62,244.46) and (326,243.84) .. (326,243.07) -- cycle ;
\draw  [fill=white] (327.9,190) .. controls (327.9,188.83) and (328.68,188.06) .. (329.63,188.06) .. controls (330.59,188.06) and (331.37,188.83) .. (331.37,189.79) .. controls (331.37,190.75) and (330.59,191.53) .. (329.63,191.53) .. controls (328.68,191.53) and (327.9,190.75) .. (327.9,189.79) -- cycle ;
\draw  [fill=white] (328.03,195.53) .. controls (328.03,194.57) and (328.81,193.79) .. (329.77,193.79) .. controls (330.72,193.79) and (331.5,194.57) .. (331.5,195.53) .. controls (331.5,196.48) and (330.72,197.26) .. (329.77,197.26) .. controls (328.81,197.26) and (328.03,196.48) .. (328.03,195.53) -- cycle ;
\draw [fill=white]  (328.03,200.99) .. controls (328.03,200.03) and (328.81,199.26) .. (329.77,199.26) .. controls (330.72,199.26) and (331.5,200.03) .. (331.5,200.99) .. controls (331.5,201.95) and (330.72,202.73) .. (329.77,202.73) .. controls (328.81,202.73) and (328.03,201.95) .. (328.03,200.99) -- cycle ;
\draw  [fill=white] (328.92,223.29) .. controls (328.92,222.98) and (329.17,222.73) .. (329.48,222.73) .. controls (329.8,222.73) and (330.05,222.98) .. (330.05,223.29) .. controls (330.05,223.6) and (329.8,223.86) .. (329.48,223.86) .. controls (329.17,223.86) and (328.92,223.6) .. (328.92,223.29) -- cycle ;
\draw [fill=white]  (328.9,226.51) .. controls (328.9,226.2) and (329.15,225.94) .. (329.47,225.94) .. controls (329.78,225.94) and (330.03,226.2) .. (330.03,226.51) .. controls (330.03,226.82) and (329.78,227.08) .. (329.47,227.08) .. controls (329.15,227.08) and (328.9,226.82) .. (328.9,226.51) -- cycle ;
\draw  [fill=white] (328.9,219.9) .. controls (328.9,219.59) and (329.15,219.33) .. (329.46,219.33) .. controls (329.78,219.33) and (330.03,219.59) .. (330.03,219.9) .. controls (330.03,220.21) and (329.78,220.47) .. (329.46,220.47) .. controls (329.15,220.47) and (328.9,220.21) .. (328.9,219.9) -- cycle ;
\draw  [fill=white] (327.9,238.99) .. controls (327.9,238.03) and (328.68,237.26) .. (329.63,237.26) .. controls (330.59,237.26) and (331.37,238.03) .. (331.37,238.99) .. controls (331.37,239.95) and (330.59,240.73) .. (329.63,240.73) .. controls (328.68,240.73) and (327.9,239.95) .. (327.9,238.99) -- cycle ;
\draw [fill=white]  (328.03,232.99) .. controls (328.03,232.03) and (328.81,231.26) .. (329.77,231.26) .. controls (330.72,231.26) and (331.5,232.03) .. (331.5,232.99) .. controls (331.5,233.95) and (330.72,234.73) .. (329.77,234.73) .. controls (328.81,234.73) and (328.03,233.95) .. (328.03,232.99) -- cycle ;
\draw [fill=white]  (327.9,207.25) .. controls (327.9,206.29) and (328.68,205.51) .. (329.63,205.51) .. controls (330.59,205.51) and (331.37,206.29) .. (331.37,207.25) .. controls (331.37,208.2) and (330.59,208.98) .. (329.63,208.98) .. controls (328.68,208.98) and (327.9,208.2) .. (327.9,207.25) -- cycle ;
\draw  [fill=white] (328.03,212.98) .. controls (328.03,212.02) and (328.81,211.21) .. (329.77,211.21) .. controls (330.72,211.21) and (331.5,212.02) .. (331.5,212.98) .. controls (331.5,213.94) and (330.72,214.71) .. (329.77,214.71) .. controls (328.81,214.71) and (328.03,213.94) .. (328.03,212.98) -- cycle ;
\draw  (438.33,154.42) .. controls (438.33,149.58) and (442.21,145.67) .. (447,145.67) .. controls (451.79,145.67) and (455.67,149.58) .. (455.67,154.42) .. controls (455.67,159.25) and (451.79,163.15) .. (447,163.15) .. controls (442.21,163.15) and (438.33,159.25) .. (438.33,154.42) -- cycle ; \draw   (438.33,154.42) -- (455.67,154.42) ; \draw   (447,145.67) -- (447,163.15) ;
\draw    (426.67,83.65) -- (447.33,83.17) -- (447.01,141.67) ;
\draw [shift={(447,143.67)}, rotate = 270.24] [color={rgb, 255:red, 0; green, 0; blue, 0 }  ][line width=0.75]    (10.93,-3.29) .. controls (6.95,-1.4) and (3.31,-0.3) .. (0,0) .. controls (3.31,0.3) and (6.95,1.4) .. (10.93,3.29)   ;
\draw    (428,222.75) -- (445.67,222.5) -- (446.97,166.17) ;
\draw [shift={(447,164.17)}, rotate = 90.92] [color={rgb, 255:red, 0; green, 0; blue, 0 }  ][line width=0.75]    (10.93,-3.29) .. controls (6.95,-1.4) and (3.31,-0.3) .. (0,0) .. controls (3.31,0.3) and (6.95,1.4) .. (10.93,3.29)   ;
\draw  [fill=blu] (483.67,103.43) .. controls (483.67,102.66) and (484.29,102.04) .. (485.05,102.04) -- (489.21,102.04) .. controls (489.97,102.04) and (490.6,102.66) .. (490.6,103.43) -- (490.6,203.45) .. controls (490.6,204.21) and (489.97,204.83) .. (489.21,204.83) -- (485.05,204.83) .. controls (484.29,204.83) and (483.67,204.21) .. (483.67,203.45) -- cycle ;
\draw   [fill=white] (485.57,106.5) .. controls (485.57,105.54) and (486.34,104.77) .. (487.3,104.77) .. controls (488.26,104.77) and (489.03,105.54) .. (489.03,106.5) .. controls (489.03,107.46) and (488.26,108.23) .. (487.3,108.23) .. controls (486.34,108.23) and (485.57,107.46) .. (485.57,106.5) -- cycle ;
\draw   [fill=white] (485.7,112.23) .. controls (485.7,111.28) and (486.48,110.5) .. (487.43,110.5) .. controls (488.39,110.5) and (489.17,111.28) .. (489.17,112.23) .. controls (489.17,113.19) and (488.39,113.97) .. (487.43,113.97) .. controls (486.48,113.97) and (485.7,113.19) .. (485.7,112.23) -- cycle ;
\draw   [fill=white] (485.7,117.7) .. controls (485.7,116.74) and (486.48,115.97) .. (487.43,115.97) .. controls (488.39,115.97) and (489.17,116.74) .. (489.17,117.7) .. controls (489.17,118.66) and (488.39,119.43) .. (487.43,119.43) .. controls (486.48,119.43) and (485.7,118.66) .. (485.7,117.7) -- cycle ;
\draw  [fill=white]  (486.58,183) .. controls (486.58,182.69) and (486.84,182.43) .. (487.15,182.43) .. controls (487.46,182.43) and (487.72,182.69) .. (487.72,183) .. controls (487.72,183.31) and (487.46,183.57) .. (487.15,183.57) .. controls (486.84,183.57) and (486.58,183.31) .. (486.58,183) -- cycle ;
\draw  [fill=white]  (486.57,186.22) .. controls (486.57,185.9) and (486.82,185.65) .. (487.13,185.65) .. controls (487.45,185.65) and (487.7,185.9) .. (487.7,186.22) .. controls (487.7,186.53) and (487.45,186.78) .. (487.13,186.78) .. controls (486.82,186.78) and (486.57,186.53) .. (486.57,186.22) -- cycle ;
\draw  [fill=white]  (486.56,179.61) .. controls (486.56,179.3) and (486.82,179.04) .. (487.13,179.04) .. controls (487.44,179.04) and (487.7,179.3) .. (487.7,179.61) .. controls (487.7,179.92) and (487.44,180.18) .. (487.13,180.18) .. controls (486.82,180.18) and (486.56,179.92) .. (486.56,179.61) -- cycle ;
\draw [fill=white]  (485.57,199.7) .. controls (485.57,198.74) and (486.34,197.97) .. (487.3,197.97) .. controls (488.26,197.97) and (489.03,198.74) .. (489.03,199.7) .. controls (489.03,200.66) and (488.26,201.43) .. (487.3,201.43) .. controls (486.34,201.43) and (485.57,200.66) .. (485.57,199.7) -- cycle ;

\draw [fill=white]  (485.7,193.7) .. controls (485.7,192.74) and (486.48,191.97) .. (487.43,191.97) .. controls (488.39,191.97) and (489.17,192.74) .. (489.17,193.7) .. controls (489.17,194.66) and (488.39,195.43) .. (487.43,195.43) .. controls (486.48,195.43) and (485.7,194.66) .. (485.7,193.7) -- cycle ;
Shape: Circle [id:dp8937358237203679] 
\draw [fill=white]  (485.57,123.95) .. controls (485.57,123) and (486.34,122.22) .. (487.3,122.22) .. controls (488.26,122.22) and (489.03,123) .. (489.03,123.95) .. controls (489.03,124.91) and (488.26,125.69) .. (487.3,125.69) .. controls (486.34,125.69) and (485.57,124.91) .. (485.57,123.95) -- cycle ;
\draw  [fill=white]  (485.7,129.69) .. controls (485.7,128.73) and (486.48,127.95) .. (487.43,127.95) .. controls (488.39,127.95) and (489.17,128.73) .. (489.17,129.69) .. controls (489.17,130.65) and (488.39,131.42) .. (487.43,131.42) .. controls (486.48,131.42) and (485.7,130.65) .. (485.7,129.69) -- cycle ;
\draw [fill=white]  (485.57,137.5) .. controls (485.57,136.54) and (486.34,135.77) .. (487.3,135.77) .. controls (488.26,135.77) and (489.03,136.54) .. (489.03,137.5) .. controls (489.03,138.46) and (488.26,139.23) .. (487.3,139.23) .. controls (486.34,139.23) and (485.57,138.46) .. (485.57,137.5) -- cycle ;
\draw [fill=white]  (485.7,143.23) .. controls (485.7,142.28) and (486.48,141.5) .. (487.43,141.5) .. controls (488.39,141.5) and (489.17,142.28) .. (489.17,143.23) .. controls (489.17,144.19) and (488.39,144.97) .. (487.43,144.97) .. controls (486.48,144.97) and (485.7,144.19) .. (485.7,143.23) -- cycle ;
\draw [fill=white]  (485.7,148.7) .. controls (485.7,147.74) and (486.48,146.97) .. (487.43,146.97) .. controls (488.39,146.97) and (489.17,147.74) .. (489.17,148.7) .. controls (489.17,149.66) and (488.39,150.43) .. (487.43,150.43) .. controls (486.48,150.43) and (485.7,149.66) .. (485.7,148.7) -- cycle ;
\draw  [fill=white] (485.57,154.95) .. controls (485.57,154) and (486.34,153.22) .. (487.3,153.22) .. controls (488.26,153.22) and (489.03,154) .. (489.03,154.95) .. controls (489.03,155.91) and (488.26,156.69) .. (487.3,156.69) .. controls (486.34,156.69) and (485.57,155.91) .. (485.57,154.95) -- cycle ;
\draw [fill=white]  (485.7,160.69) .. controls (485.7,159.73) and (486.48,158.95) .. (487.43,158.95) .. controls (488.39,158.95) and (489.17,159.73) .. (489.17,160.69) .. controls (489.17,161.65) and (488.39,162.42) .. (487.43,162.42) .. controls (486.48,162.42) and (485.7,161.65) .. (485.7,160.69) -- cycle ;
\draw  [fill=white] (485.57,173.7) .. controls (485.57,172.74) and (486.34,171.97) .. (487.3,171.97) .. controls (488.26,171.97) and (489.03,172.74) .. (489.03,173.7) .. controls (489.03,174.66) and (488.26,175.43) .. (487.3,175.43) .. controls (486.34,175.43) and (485.57,174.66) .. (485.57,173.7) -- cycle ;
\draw [fill=white]  (485.7,167.7) .. controls (485.7,166.74) and (486.48,165.97) .. (487.43,165.97) .. controls (488.39,165.97) and (489.17,166.74) .. (489.17,167.7) .. controls (489.17,168.66) and (488.39,169.43) .. (487.43,169.43) .. controls (486.48,169.43) and (485.7,168.66) .. (485.7,167.7) -- cycle ;

\draw    (455.67,154) -- (480.5,154) ;
\draw [shift={(482.5,154)}, rotate = 180] [color={rgb, 255:red, 0; green, 0; blue, 0 }  ][line width=0.75]    (10.93,-3.29) .. controls (6.95,-1.4) and (3.31,-0.3) .. (0,0) .. controls (3.31,0.3) and (6.95,1.4) .. (10.93,3.29)   ;

\draw  [dash pattern={on 4.5pt off 4.5pt}] (358,56.2) -- (465,56.2) -- (465,246) -- (358,246) -- cycle ;
\draw [fill=blu]  (530,131) -- (535,131) -- (535,175.17) -- (530,175.17) -- cycle ;

\draw [fill=blu]  (510,131) -- (515,131) -- (515,175.17) -- (510,175.17) -- cycle ;
\draw    (480,154) -- (530,154) ;
\draw [shift={(530,154)}, rotate = 180] [color={rgb, 255:red, 0; green, 0; blue, 0 }  ][line width=0.75]    (10.93,-3.29) .. controls (6.95,-1.4) and (3.31,-0.3) .. (0,0) .. controls (3.31,0.3) and (6.95,1.4) .. (10.93,3.29)   ;
\draw    (535,139) -- (550,139) ;
\draw [shift={(550.33,139)}, rotate = 180] [color={rgb, 255:red, 0; green, 0; blue, 0 }  ][line width=0.75]    (6.56,-1.97) .. controls (4.17,-0.84) and (1.99,-0.18) .. (0,0) .. controls (1.99,0.18) and (4.17,0.84) .. (6.56,1.97)   ;
\draw    (535,154) -- (550,154) ;
\draw [shift={(550,154)}, rotate = 180] [color={rgb, 255:red, 0; green, 0; blue, 0 }  ][line width=0.75]    (6.56,-1.97) .. controls (4.17,-0.84) and (1.99,-0.18) .. (0,0) .. controls (1.99,0.18) and (4.17,0.84) .. (6.56,1.97)   ;
\draw    (535,171) -- (550,171) ;
\draw [shift={(550,171)}, rotate = 180] [color={rgb, 255:red, 0; green, 0; blue, 0 }  ][line width=0.75]    (6.56,-1.97) .. controls (4.17,-0.84) and (1.99,-0.18) .. (0,0) .. controls (1.99,0.18) and (4.17,0.84) .. (6.56,1.97)   ;
\draw (415.01,50) node  [font=\footnotesize] [align=left] {\begin{minipage}[lt]{150.11pt}\setlength\topsep{0pt}
\begin{center}
\textbf{Modality Fusion}
\end{center}

\end{minipage}};
\draw (370,185) node [anchor=north west][inner sep=0.75pt]  [font=\footnotesize] [align=left] {\begin{minipage}[lt]{69.58pt}\setlength\topsep{0pt}

\end{minipage}};
\draw (435.67,132) node [anchor=north west][inner sep=0.75pt]  [font=\footnotesize] [align=left] {\begin{minipage}[lt]{57.48pt}\setlength\topsep{0pt}

\end{minipage}};
\draw (439,195) node [anchor=north west][inner sep=0.75pt]  [font=\footnotesize] [align=left] {\begin{minipage}[lt]{60.12pt}\setlength\topsep{0pt}

\end{minipage}};
\draw [fill=grn]  (280,70) -- (285,70) -- (285,110) -- (280,110) -- cycle ;
\draw (263,110) node [anchor=north west][inner sep=0.15pt]  [font=\small] [align=center,font=\tiny\linespread{0.8}\selectfont] {Dropout \& \\  Normalise};
\draw [fill=yel]  (280,195) -- (285,195) -- (285,235) -- (280,235) -- cycle ;
\draw (263,177) node [anchor=north west][inner sep=0.15pt]  [font=\small] [align=center,font=\tiny\linespread{0.8}\selectfont] {Dropout \& \\  Normalise};
\draw (370,91) node [anchor=north west][inner sep=0.75pt]  [font=\footnotesize] [align=left] {\begin{minipage}[lt]{69.58pt}\setlength\topsep{0pt}

\end{minipage}};

\draw (305,30) node [anchor=north west][inner sep=0.75pt]  [font=\footnotesize] [align=left] {\begin{minipage}[lt]{25.56pt}\setlength\topsep{0pt}
\begin{center}
{\normalsize Visual}\\{\normalsize Features}
\end{center}

\end{minipage}};
\draw (305,150.67) node [anchor=north west][inner sep=0.75pt]  [font=\footnotesize] [align=left] {\begin{minipage}[lt]{25.56pt}\setlength\topsep{0pt}
\begin{center}
{\normalsize Textual}\\{\normalsize Features}
\end{center}

\end{minipage}};
\draw (470,60) node [anchor=north west][inner sep=0.75pt]  [font=\footnotesize] [align=left] {\begin{minipage}[lt]{60.31pt}\setlength\topsep{0pt}
{\normalsize{Fused\\Multimodal Features}}

\end{minipage}};

\draw (493,175) node [anchor=north west][inner sep=0.15pt]  [font=\small] [align=center,font=\tiny\linespread{0.8}\selectfont] {Dropout \\ \& \\  Normalise};

\draw (567,127) node [anchor=north west][inner sep=0.75pt][rotate=270] [font=\small] [align=center] {Sentiment};
\draw    (90,215) -- (144,215) ;
\draw [shift={(144,215)}, rotate = 180] [color={rgb, 255:red, 0; green, 0; blue, 0 }  ][line width=0.75]    (10.93,-3.29) .. controls (6.95,-1.4) and (3.31,-0.3) .. (0,0) .. controls (3.31,0.3) and (6.95,1.4) .. (10.93,3.29)   ;
\draw    (255,215) -- (320,215) ;
\draw [shift={(320,215)}, rotate = 180] [color={rgb, 255:red, 0; green, 0; blue, 0 }  ][line width=0.75]    (10.93,-3.29) .. controls (6.95,-1.4) and (3.31,-0.3) .. (0,0) .. controls (3.31,0.3) and (6.95,1.4) .. (10.93,3.29)   ;

\draw    (255,90) -- (320,90) ;
\draw [shift={(320,90)}, rotate = 180] [color={rgb, 255:red, 0; green, 0; blue, 0 }  ][line width=0.75]    (10.93,-3.29) .. controls (6.95,-1.4) and (3.31,-0.3) .. (0,0) .. controls (3.31,0.3) and (6.95,1.4) .. (10.93,3.29)   ;

\draw    (90,90) -- (144,90) ;
\draw [shift={(144,90)}, rotate = 180] [color={rgb, 255:red, 0; green, 0; blue, 0 }  ][line width=0.75]    (10.93,-3.29) .. controls (6.95,-1.4) and (3.31,-0.3) .. (0,0) .. controls (3.31,0.3) and (6.95,1.4) .. (10.93,3.29)   ;

\draw  [fill=grn]  (144,67) -- (255,67) -- (255,113) -- (144,113) -- cycle  ;
\draw (200,89) node   [align=left] {\begin{minipage}[lt]{71.9pt}\setlength\topsep{0pt}
\begin{center}
CLIP\\Image Encoder
\end{center}

\end{minipage}};
\draw  [fill=yel]  (144,192) -- (255,192) -- (255,238) -- (144,238) -- cycle  ;
\draw (200,215) node   [align=left] {\begin{minipage}[lt]{62.25pt}\setlength\topsep{0pt}
\begin{center}
CLIP\\Text Encoder
\end{center}

\end{minipage}};

\draw (60,185) node [anchor=north west][inner sep=0.75pt]  [font=\scriptsize] [align=left] {\begin{minipage}[lt]{72.52pt}\setlength\topsep{0pt}
\begin{center}
{\normalsize Text\\Input}
\end{center}

\end{minipage}};

\draw (60,60) node [anchor=north west][inner sep=0.75pt]  [font=\scriptsize] [align=left] {\begin{minipage}[lt]{72.52pt}\setlength\topsep{0pt}
\begin{center}
{\normalsize Image\\Input}
\end{center}

\end{minipage}};

\end{tikzpicture}